\documentclass[accepted]{uai2023} 

\usepackage{microtype}
\usepackage{graphicx}
\usepackage{subfigure}
\usepackage{booktabs} 
\usepackage{hyperref}

\usepackage{amsmath}
\usepackage{amssymb}
\usepackage{mathtools}
\usepackage{amsthm}
\usepackage[capitalize,noabbrev]{cleveref}
\theoremstyle{plain}
\newtheorem{theorem}{Theorem}[section]
\newtheorem{proposition}[theorem]{Proposition}

\theoremstyle{definition}
\newtheorem{definition}[theorem]{Definition}
\newtheorem{assumption}[theorem]{Assumption}
\theoremstyle{remark}
\newtheorem{remark}[theorem]{Remark}
\usepackage[textsize=tiny]{todonotes}
\usepackage{dsfont}
\usepackage{xr} 
\usepackage[american]{babel}

\usepackage{natbib} 
\usepackage{mathtools} 
\usepackage{booktabs} 
\usepackage{tikz} 

\title{Efficient Failure Pattern Identification of Predictive Algorithms}

\author[1,2]{Bao Nguyen}
\author[3]{Viet Anh Nguyen}

\affil[1]{%
    School of Information and Communication Technology\\
    Hanoi University of Science and Technology\\
    Vietnam
}
\affil[2]{%
    College of Engineering \& Computer Science\\
    VinUni-Illinois Smart Health Center\\
    VinUniversity\\
    Vietnam
}
\affil[3]{%
    The Chinese University of Hong Kong
}
\makeatletter
\def\munderbar#1{\underline{\sbox\tw@{$#1$}\dp\tw@\z@\box\tw@}}
\makeatother

\newcommand{\be}{\begin{equation}}
\newcommand{\ee}{\end{equation}}
\newcommand{\bea}{\begin{equation*}\begin{aligned}}
\newcommand{\eea}{\end{aligned}\end{equation*}}

\newcommand{\R}{\mathbb{R}}

\newcommand{\wh}{\widehat}
\newcommand{\mc}{\mathcal}

\newcommand{\cov}{\Sigma} 
\newcommand{\covsa}{\wh{\cov}}

\newcommand{\PSD}{\mathbb{S}_{+}} 
\newcommand{\Let}{\triangleq}

\newcommand{\EE}{\mathds{E}}

\newcommand{\msa}{\wh \mu}

\newcommand{\true}{\mathrm{true}}
\newcommand{\VoI}{\mathrm{VoI}}
\begin{document}
\maketitle

\begin{abstract}
    Given a (machine learning) classifier and a collection of unlabeled data, how can we efficiently identify misclassification patterns presented in this dataset? To address this problem, we propose a human-machine collaborative framework that consists of a team of human annotators and a sequential recommendation algorithm. The recommendation algorithm is conceptualized as a stochastic sampler that, in each round, queries the annotators a subset of samples for their true labels and obtains the feedback information on whether the samples are misclassified. The sampling mechanism needs to balance between discovering new patterns of misclassification (exploration) and confirming the potential patterns of classification (exploitation). We construct a determinantal point process, whose intensity balances the exploration-exploitation trade-off through the weighted update of the posterior at each round to form the generator of the stochastic sampler. The numerical results empirically demonstrate the competitive performance of our framework on multiple datasets at various signal-to-noise ratios.
\end{abstract}

\section{Introduction} 
\label{sec:intro}


Over the past few years, algorithmic predictive models have claimed many successful stories in real-world applications, ranging from healthcare and finance to jurisdiction and autonomous driving. These successes often take place in an invariant environment where the training data and the test data come from sufficiently similar distributions. If this similarity condition does not hold, then it is well-known that the performance of the algorithmic prediction can deteriorate significantly in the deployment phase. This performance deterioration may trigger subsequent concerns, especially in consequential domains such as self-driving cars and healthcare, where the algorithmic predictions may affect system reliability and human safety. When a predictive model performs unsatisfactorily in a \textit{systematic} manner, then it is called a failure pattern. For example, if an object detection system fails to capture the object systematically under low-light conditions, then it is a failure pattern. Detecting and correcting the failure patterns is arguably one of the most daunting challenges in developing the future analytics systems.

Detecting failure patterns is also beneficial for many steps in the life-cycle of an analytics product. For example, it is very common to develop analytics systems using data collected from one country, say the United States, but the systems can be deployed in another country, say Canada. Before field deployment, the systems need to undergo intensive fine-tuning and recalibration, and information regarding the failures can significantly reduce the time and efforts spent on this step. Further, as foundation models are increasingly used as a building block of future analytics systems, assessing the blindspots of these pre-trained models is of crucial importance for product development, deployment and continuous performance evaluation.

Detecting failure patterns, unfortunately, requires the true outcome or label of the data samples. If the dataset is cleanly labeled, then the problem of failure pattern detection can be simplified to a search problem. The situation becomes more cumbersome if we have access to only an \textit{un}labeled dataset. It is thus natural herein to incorporate the role of a human annotator into the detection procedure. However, in many applications that require high-skilled annotators, such as healthcare, it is time-consuming and cost-intensive to query the annotator. Given a dataset containing \textit{un}labeled samples, we are interested in designing an efficient routine to query these samples for their true outcome or label, so that we can identify as many failure patterns as possible with minimal annotation queries.

\textbf{Contributions.} We propose in this paper a directed sampling mechanism with the goal of detecting failure patterns from an unlabeled dataset. This mechanism has two main components:
\begin{itemize}[leftmargin=3mm]
    \item a Gaussian process to model the predictive belief about the misclassification probability for each unlabeled sample,
    \item a determinantal point process sampling that balances the trade-off between exploration and exploitation by taking a mixture between a similarity matrix (reflecting exploration) and a posterior probability of misclassification matrix (reflection exploitation).
\end{itemize}

Ultimately, we propose a human-machine collaborative framework that relies on the machine's proposed queries and the corresponding annotator's feedback to detect failure patterns, and consequently improve the reliability of the predictive algorithm in variant environments.

This paper unfolds as follows: In Section~\ref{sec:lit}, we survey existing efforts to identify failure patterns in a given dataset. Section~\ref{sec:problem} formalizes our problem and introduces the necessary notations. Section~\ref{sec:GP} depicts our approach using Gaussian processes to build the surrogate for the Value-of-Interest, which represents the belief about the misclassification probability for the unlabeled samples. Section~\ref{sec:dpp} focuses on the determinantal point process sampler, which blends the Value-of-Interest (highlighting exploitation) with a diversity term (highlighting exploration). Section~\ref{sec:bandwidth} discusses how we can choose the bandwidth, which is critical given the unsupervised nature of the failure identification problem. Finally, Section~\ref{sec:experiment} provides extensive numerical results to demonstrate the superior performance of our directed sampling mechanism in detecting failure patterns for various datasets.

In almost all related work about failure identification, a failure mode (termed as ``slice'' in~\citet{ref:eyuboglu2022domino}) of a pre-trained classifier on a dataset is a subset of the dataset that meets two conditions: (i) the classifier performs poorly when predicting samples in that subset, and (ii) the subset captures a distinct concept or attribute that would be recognizable to domain experts. If the true labels of all samples in that subset are available, criterion (i) can be easily confirmed using popular performance metrics for classification tasks. However, condition (ii) is subjective and implicit. For instance, two medical experts may interpret a group of misclassified brain scan images in two different ways, and both interpretations may be reasonable and acceptable. For unlabeled data, it is difficult to employ the existing definitions of the failure patterns.

Arguably, a failure mode is a subjective term that depends on the machine learning task, on the users, and on the datasets. We do not aim to provide a normative answer to the definition of a failure mode in this paper. Our paper takes a pragmatic approach: given a choice of failure pattern definition of users, we develop a reasonable method that can efficiently find the failure patterns from the unlabeled data.

\textbf{Notations. } We use $\R^d$ to denote the space of $d$-dimensional vectors, and $\PSD^d$ to denote the set of $d$-by-$d$ symmetric, positive semidefinite matrices. The transposition of a matrix $A$ is denoted by $A^\top$, and the Frobenius norm of a matrix $A$ is denoted by $\| A \|_F$.

\section{Related work} \label{sec:lit}

Detecting failure patterns of a machine learning algorithm on a given dataset is an emergent problem in the literature. \citet{ref:deon2022spotlight} formulates a single-stage optimization problem to identify the systematic error.  \citet{ref:sohoni2020no} mitigates hidden stratification by detecting sub-class labels before solving a group distributionally robust optimization problem(GDRO). \citet{ref:eyuboglu2022domino} provides an evaluation method for slice discovery methods (SDMs): given a trained classifier and a labeled dataset of the form $(x_i, y_i)_{i=1}^N$, it outputs a set of slicing functions that partitions the dataset into overlapping subgroups on which the model underperforms. \citet{nushi2018towards} proposes a hybrid approach to analyze the failures of AI systems in a more accountable and transparent way. The hybrid approach involves two main phases: (1) machine analysis to identify potential failure modes, and (2) human analysis to validate and explain the identified failure modes.  \citet{singla2021understanding} concentrates on understanding the failures of deep neural networks. The paper proposes a method for robust feature extraction, which aims to identify the most important features that contribute to the output of a deep neural network. \citet{47966} proposed SliceFinder that automatically identifies subsets of the data that cause the model to fail or perform poorly. It repeatedly draws random subsets of the data and re-trains the model on each subset, then measures the model's performance. Finally, it employs a statistical test to identify data slices that are significantly different from the overall distribution. \citet{sagadeeva2021sliceline} build a method using linear algebra techniques to speed up the slice-finding process for identifying problematic subsets of data that cause a model to fail.

Our paper is also closely related to the field of active learning. Active learning examines how to obtain the largest amount of performance gains by labeling as few samples as possible. Comprehensive surveys on active learning can be found in \citet{ref:ren2021survey} and \citet{ref:budd2021survey}. There is, however, a critical difference between the settings of our paper and those of active learning: in our paper, we focus on identifying the failure pattern for a fixed (invariant) classifier, whereas active learning focuses on improving the model performance with model retraining after each batch of recommendations. The numerical experiments in Section~\ref{sec:experiment}
 demonstrate empirically that active learning algorithms do not perform competitively at the failure identification task.

\begin{figure}
    \centering
    \includegraphics[width=\linewidth]{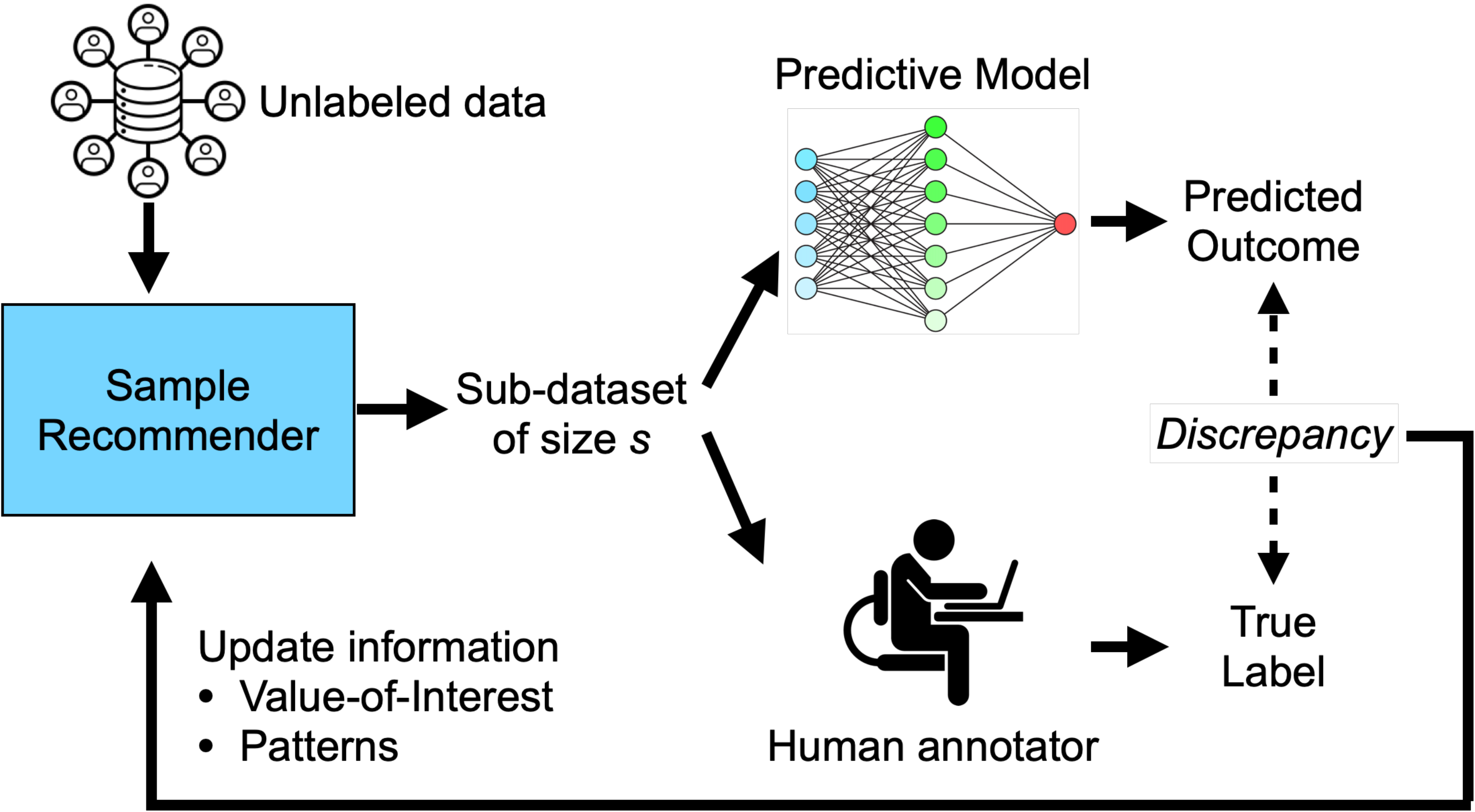}
    \caption{Schematic view of our solution: We build a sampling mechanism to recommend the next $s$ unlabeled samples to be labeled by the annotators. The misclassification information is fed back to update the sampling density. Throughout this sequential recommendation process, the predictive model does not change.}
    \label{fig:sketch}
\end{figure}

\section{Problem Statement} \label{sec:problem}

We suppose that a user is given a classification algorithm and a set of unlabeled dataset. The classifier is denoted by $\mc C: \mc X \to \mc Y$, which admits a feature space $\mc X = \R^d$ as input, and outputs labels in the set $\mc Y = \{1, \ldots, C\}$. The unlabeled dataset consists of $N$ samples, which can be represented by $N$ vectors $x_i \in \R^d$ for $i = 1, \ldots, N$. For each sample $x_i$, the predicted label (termed the pseudolabel) that is recommended by the algorithm is denoted by $\wh y_i = \mc C(x_i)$; while its true label is denoted by $y_i^{\true}$. The $i$-th sample in the dataset is accurately classified if $\wh y_i = y_i^{\true}$, and it is misclassified if $\wh y_i \neq y_i^{\true}$.

The main object of this paper is \textit{not} the individual misclassified samples. Our goal in this paper is to study the group effect of misclassified samples: when they are sufficiently close to each other, they form a failure pattern or a failure cluster, and together, they signal a systematic error of the predictive algorithm on the given dataset. To give a formal definition of a failure pattern, we need to construct a graph representation of the data. To do this, we suppose the available data can form an abstract undirected graph $\mc G = (\mc V, \mc E)$, where $\mc V$ is the set of nodes whereas each node represents one sample from the dataset, and $\mc E$ is the set of edges. Let $\mc G^{\mathrm{mis}} = (\mc V^{\mathrm{mis}}, \mc E^{\mathrm{mis}})$ be the pruned graph of $\mc G$ after removing the nodes representing samples that are classified accurately. We can now rigorously define a general failure pattern.

\begin{definition}[$M$-C failure pattern]
    \label{def:pattern}
     Given a graph $\mathcal G^{\mathrm{mis}}$, an integer size threshold $M$, and a connectivity criterion C, a failure pattern is a subgraph $\mathcal G^{\mathrm{fail}} = (\mathcal V^{\mathrm{fail}}, \mathcal E^{\mathrm{fail}})$ of $\mathcal G^{\mathrm{mis}}$ that satisfies the connectivity criterion C and the cardinality of the set $\mathcal V^{\mathrm{fail}}$ is at least $M$.
\end{definition}

Our framework gives the user tremendous flexibility in specifying the problem and leveraging our tool to identify the related failure patterns: 
\begin{itemize}[leftmargin=5mm]
\item First, the user can specify the semantic similarity between two nodes represented by an edge connecting them. Moreover, this user-based graph construction can capture the inherently subjective characteristics of the definition of failure, which depends on the user's perspective and the domain of data. For a concrete example, if the user is confident that the embedding space is rich enough to capture the similarity between samples, then one possible approach is to construct $\mc G$ by the mutually nearest neighbor graph under the Euclidean distance between embedding representations. 
\item Secondly, users can specify preferred connectivity criteria among samples in a failure pattern. For instance, the criterion of ``completeness'' can be employed as a candidate for C. This criterion is one of the most stringent: it mandates that each misclassified node must exhibit an edge (indicating semantic similarity) with every other misclassified node in the pattern. However, users can adjust the stringency by employing looser criteria, such as ``Super-$\kappa$ connectivity'' or ``connected connectivity''. 
\item Third, the users can specify $M$ which depicts the amount of evidence required in order to confirm a failure pattern. A large value of $M$ means a high level of evidence required (represented by a high number of clustered misclassified samples) to pinpoint a pattern. Moreover, one can also see that given a fixed graph of misclassified samples $\mc G^{\mathrm{mis}}$, the number of failure patterns in $\mc G^{\mathrm{mis}}$ is non-increasing in $M$. Taking this perspective, one can also view $M$ as a parameter that is \textit{inverse}-proportional to the user's degree of risk aversion: a low value of $M$ indicates that the user believes there are many misclassified patterns in the dataset.
\end{itemize}


While identifying the failure patterns from an unlabeled dataset is becoming increasingly pertinent in the current practice of machine learning deployment, this problem is inherently challenging due to two main factors:
    \begin{enumerate}[leftmargin=5mm]
        \item Annotation cost: to determine if a data point is misclassified, we need to have complete information about its \textit{true} label, which is often obtained by querying a team of human annotators. Unfortunately, in many consequential domains such as healthcare and law enforcement, the cost of annotation can be enormous for a large dataset.
        \item Signal-to-noise ratio: suppose that there are $F$ failure patterns represented in the data graph $\mc G^{\mathrm{mis}}$ and they are denoted by subgraphs $\mc G^{\mathrm{fail}}_f =  (\mc V^{\mathrm{fail}}_f, \mc E^{\mathrm{fail}}_f)$ for $f = 1, \ldots, F$. The union $\cup_{f} \mc V^{\mathrm{fail}}_f$ gives a collection of all misclassifed samples that belong to the failure patterns, and the remainder set $\mc V^{\mathrm{mis}} \backslash \cup_{f} \mc V^{\mathrm{fail}}_f$ contains all samples that are misclassified but do not belong to any failure pattern. Because the goal of the user is to identify failure patterns, the user can view the cardinality of the union set $\cup_{f} \mc V^{\mathrm{fail}}_f$ as a measure of the amount of signal in the dataset, and the cardinality of the remainder set $\mc V^{\mathrm{mis}} \backslash \cup_{f} \mc V^{\mathrm{fail}}_f$ as a measure of the amount of noise therein. As such, we observe a typical signal-to-noise ratio (SNR) indication: if the SNR is high, then the problem tends to be easy; on the contrary, if the SNR is low then the problem tends to be hard.
    \end{enumerate}

Unfortunately, crucial information to identify the failure patterns is not known to the user ex-ante: for example, the user does not know how many patterns there are in the dataset, nor does the user know the number of misclassified samples and the SNR. To proceed, we make the assumptions:
\begin{assumption}
    We assume the followings:
    \begin{itemize}[leftmargin=5mm]
        \item the number of unlabeled samples $N$ in the given dataset is large and the cost of using an annotator is expensive, thus it may be prohibitive to annotate the whole dataset,
        \item the inference cost of the predictive algorithm is negligible for any feature $x \in \mc X$, which means we can obtain the pseudolabels $\wh y_i = \mc C(x_i)$ for all data points,
        \item  the predictive algorithm $\mathcal{C}$ remains invariant throughout the failure identification process, and thus the failure patterns do not change over time.
    \end{itemize}
\end{assumption}

The first assumption is critical because if the cost of annotating the whole dataset is small, then the user does not need to use our proposed sampling mechanism because the benefit of annotating the whole dataset outweighs the resulting cost. The second assumption is reasonable in most practical settings as it requires feed-forwarding the entire given dataset through the predictive algorithm once. The last assumption ensures that our targets, the failure patterns, do not alter over time and that the collected information and the updated belief remain relevant for recommending the next batch of samples for annotation. 

We are interested in a dynamic (sequential) recommendation setting that consists of $T$ rounds. In each round $t = 1, \ldots, T$, the system recommends to the user a set of $s$ unlabeled samples to query the annotator for the true label. After the true labels are obtained, the user can recognize which newly-annotated samples are misclassified, and the user, based on this arrival of information, can (i) identify if a new failure pattern can be identified by verifying the conditions in Definition~\ref{def:pattern}, and then (ii) update the user's own belief about where the remaining failure pattern may locate. The posterior belief is internalized to the recommendation mechanism to suggest another set of $s$ unlabeled samples from the remaining pool to the next round.

We note that task (i) described above is purely algorithmic: after the arrival of the misclassification information, the user can run an algorithm to search for the newly-formed maximally connected subgraphs of misclassified samples with size at least $M$. Task (ii), on the contrary, is more intricate because it requires a dynamic update of belief. To achieve task (ii), we will employ a Bayesian approach with Gaussian processes to form a belief about the locations of the failure patterns, and this belief will also be used for the sampling mechanism.

\begin{remark}[Semantics of the failure pattern] \label{rem:semantics}
Herein, the failure patterns are defined using the closeness in the feature of the samples. This is a convenient abstraction to alleviate the dependence of the definition on specific tasks (such as image, text, or speech classification). Defining patterns based on the feature space is also a common practice in failure identification, see \citet{ref:eyuboglu2022domino} and \citet{ref:deon2022spotlight}.
\end{remark}

\section{Gaussian Process for  the Value-of-Interest}
\label{sec:GP}

In this section, we describe our construction of a surrogate function called the Value-of-Interest (VoI). Mathematically, we define $\mathrm{VoI}: \mc X \times \mc Y \to [0, 1]$ to quantify the degree of interest assigned to each pair of feature-pseudolabel $(x_i, \wh y_i)$ data point. The notion of $\VoI$ aims to capture the exploitation procedure: it emphasizes recommending samples with a high tendency (or intensity) to confirm a misclassification pattern. Thinking in this way, we aim to predict the probability that the feature-pseudolabel pair will be part of a yet-to-confirmed failure pattern. For any generic sample $(x, \hat y)$, we model $\VoI$ using a sigmoid function of the form
    \[
        \VoI (x, \hat{y}) = \frac{1}{1 + \exp(- g(x, \hat y))},
    \]
    for some function $g: \mc X \times \mc Y \to \R$. While $\VoI$ is bounded between 0 and 1, the function $g$ can be unbounded, and it is more suitable to model our belief about $g$ using a Gaussian process (GP). In particular, we let $g \sim \mathrm{GP}(m,~ \mc K)$, where $m$ is the mean function and $\mc K$ is the kernel or covariance function, both are defined on $\mc X \times \mc Y$. A GP enables us to model the neighborhood influence among different samples through the covariance function $\mc K$.

\subsection{Posterior Update of the Predictive Probability}

Using the Gaussian process model, we have
\[
    \tilde g = \begin{bmatrix}
        g(x_1, \hat y_1) \\
        \vdots \\
        g(x_N, \hat y_N)
    \end{bmatrix} \sim \mathcal N (0, K),
\]
where for a slight abuse of notation, $m$ is a vector of mean values and the covariance matrix $K \in \PSD^N$ is the Gram matrix with the components
\be \label{eq:Gram}
    K_{ij} = \mathcal K((x_i, \hat y_i), (x_j, \hat y_j)) \quad \forall (i,j).
\ee
We can re-arrange $g$ into another vector $(\tilde g_{[t]}, \tilde g_{[t]}^*)$ where the subvector $\tilde g_{[t]} = (g(x_i, \hat y_i))_{i \in \mathcal I_{[t]}}$ represents the value of $\tilde g$ at all samples that are queried by time $t$, while $\tilde g_{[t]}^* = (g(x_i, \hat y_i))_{i \in \mathcal I_{[t]}^*} $ represents the value of $\tilde g$ at all samples that are \textit{not} queried yet by time $t$. By a similar decomposition of the matrix $K$, we can rewrite
\[
    \begin{bmatrix}
    \tilde g_{[t]} \\ \tilde g_{[t]}^*
    \end{bmatrix} \sim \mathcal N\left( \begin{pmatrix} m_{[t]} \\ m_{[t]}^*\end{pmatrix}, \begin{bmatrix} K_{[t]} & K_{[t]}^* \\ (K_{[t]}^*)^\top & K_{[t]}^{**} \end{bmatrix} \right).
\]
By the law of conditional distribution for joint Gaussian distributions~\citep[\S15.2.1]{ref:murphy2012machine}, we have the distribution of $\tilde g_{[t]}^*$ conditional on $\tilde g_{[t]} =  g_{[t]}^{\text{observe}}$ is also a Gaussian distribution:
\[
    \tilde g_{[t]}^* | g_{[t]}^{\text{observe}} \sim \mathcal N(m_t^*, \cov_t^*),
\]
where the conditional mean is determined by
\begin{subequations} \label{eq:update}
\begin{align}
\label{eq:update-mean}
    m_t^*  &=  m_{[t]}^* + (K_{[t]}^*)^\top K_{[t]}^{-1} (g_{[t]}^{\text{observe}} - m_{[t]})
\end{align}
and the conditional variance is computed as
\begin{align}
\label{eq:update-cov}
    \cov_t^* &= K_{[t]}^{**} - (K_{[t]}^{*})^\top K_{[t]}^{-1} K_{[t]}^{*}.
\end{align}
\end{subequations}
The vector $m_t^*$ captures the posterior mean of the misclassification probability of samples that are not yet queried by time $t$.

We now discuss how to estimate the expected value of $\VoI(x_i, \hat y_i)$ for the \textit{un}queried sample $i$. Note that $\VoI(x_i, \hat y_i)$ is a nonlinear function of $g(x_i, \hat y_i)$, we can use a second-order Taylor expansion of $\VoI$ around the conditional mean of $g$, and we obtain\footnote{See Appendix~\ref{sec:taylor_expansion} for detailed derivation.}
\begin{align} \label{eq:mean}
    \EE[\VoI(x_i, \hat y_i) | g_{[t]}^{\text{observe}}] \approx
    \alpha_i + \frac{1}{2} \cov_{t, i}^* \beta_i \Let \gamma_{t, i},
\end{align}
with $\alpha_i$ and $\beta_i$ being computed as
\[
\alpha_i = (1 + \exp(- m_{t,i}^*))^{-1}, \quad
\beta_i = \alpha_i (1 - \alpha_i) (1 - 2 \alpha_i).
\]
In the above formulas, $m_{t,i}^*$ and $\cov_{t, i}^*$ are the mean and the variance component of the vector $m_{t}^*$ and matrix $\cov_{t}^*$ corresponding to sample $i$. A disadvantage of the approximation~\eqref{eq:mean} is that the value $\gamma_{t,i}$ may become negative due to the possible negative value of $\beta_i$. If this happens, we can resort to the first-order approximation:
\[
\EE[\VoI(x_i, \hat y_i) | g_{[t]}^{\text{observe}}] \approx
    \alpha_i,
    \]
which guarantees positivity due to the definition of $\alpha_i$.

\subsection{Covariance Specification.} 
Given any two samples $(x, \wh y)$ and $(x', \wh y')$, the covariance between $g(x, \wh y)$ and $g(x', \wh y')$ is dictated by
    \[
    \mathrm{Cov}(g(x, \wh y), g(x', \wh y')) = \mc K((x, \wh y), (x', \wh y')).
    \]
    The bivariate function $\mc K$ is constructed using a kernel on the feature-pseudolabel space $\mc X \times \mc Y$. We impose a product kernel on $\mc X \times \mc Y$ of the form
\be \label{eq:K}
    \mathcal K((x, \wh y), (x', \wh y')) = \mathcal K_{\mathcal X}(x, x') \mathcal K_{\mathcal Y}(\wh y, \wh y').
\ee
In order to express a covariance function, we construct $\mc K$ as a positive definite kernel.
\begin{definition}[Positive definite (pd) kernel] \label{def:pd-kernel}
            Let $\mc Z$ be any set. A symmetric function $\mathcal K_{\mc Z}: \mc Z \times \mc Z \to \R$ is positive definite if for any natural number $n$ and any choices 
   of $(z_i)_{i=1}^n \in \mc Z$ and $(\alpha_i)_{i=1}^n \in \R$, we have
    \[
        \sum_{i=1}^n\sum_{j=1}^n \alpha_i \alpha_j \mathcal K_{\mc Z}(z_i, z_j) \ge 0.
    \]
    \end{definition}
If $\mathcal K_{\mathcal X}$ and $\mathcal K_{\mathcal Y}$ are positive definite kernels, then $\mathcal K$ is also a positive definite kernel according to Schur product theorem \citep[Theorem~VII]{Schur+1911+1+28}. Thus, it now suffices to construct individual pd kernel for $\mc K_{\mc X}$ and $\mc K_{\mc Y}$. We choose $\mc K_{\mc X}$ as the Gaussian kernel
    \be \label{eq:Kx}
        \mc K_{\mc X}(x, x') = \exp\big( -\frac{1}{2 h_{\mc X}^2} \| x - x' \|_2^2 \big),
    \ee
    where $h_{\mc X} > 0$ is the kernel width.

The main difficulty encountered when specifying the kernel is the categorical nature of the pseudolabel. 
Imposing a kernel on $\mc Y$ hence may require a significant effort to pin down a similarity value between any pair of labels. To alleviate this difficulty, we first represent each label $y$ by the respective conditional first- and second-moments, in which the moments are estimated using the samples and their pseudolabels. More specifically, we collect all samples whose pseudolabel is $y$, and we estimate the feature mean vector and the feature covariance matrix as follows
    \begin{align*}
        & \msa_y = \frac{1}{N_{y}} \sum_{i: \hat{y}_i = y} x_i \in \R^d, \quad \text{and} \\
        & \covsa_y = \frac{1}{N_{y}} \sum_{i: \hat{y}_i = y} (x_i - \msa_y)(x_i - \msa_y)^\top \in \PSD^d,
    \end{align*}
    where $N_y$ is the number of samples with pseudolabel $y$. We now anchor the kernel on $\mc Y$ through the Gaussian kernel on the product space $\R^d \times \PSD^d$ of the mean-covariance representation, that is, for any generic label $y$ and $y'$:
    \be \label{eq:Ky}
        \mc K_{\mc Y}(y, y') = \exp\big( -\frac{\| \msa_y - \msa_{y'} \|_2^2}{2 h_{\mc Y}^2}  \big)\exp\big( -\frac{\| \covsa_y - \covsa_{y'} \|_F^2}{2 h_{\mc Y}^2}  \big).
    \ee
    Notice that this kernel is specified by a single bandwidth parameter $h_{\mc Y} > 0$. By combining \citet[Theorems~4.3 and 4.4]{ref:jayasumana2013kernel} and by the fact that the product of pd kernels is again a pd kernel, we conclude that the kernel $\mathcal{K}$ defined using the formulas~\eqref{eq:K}-\eqref{eq:Ky} is a pd kernel.

\begin{remark}[Feature-label metric]
Measuring the distance between two (categorical) classes by measuring the distance between the corresponding class-conditional distributions was previously used in~\citet{ref:alvarez2020geometric} and~\citet{ref:hua2023dynamic}. Under the Gaussian assumptions therein, the 2-Wasserstein distance between conditional distributions simplifies to an explicit formula involving a Euclidean distance between their mean vectors and a Bures distance between their covariance matrices. Unfortunately, constructing a Gaussian kernel with the Bures distance on the space of symmetric positive semidefinite matrices may not lead to a pd kernel. To ensure that $\mc K_{\mc Y}$ is a pd kernel, we need to use the Frobenius norm for the covariance part of~\eqref{eq:Ky}. 
\end{remark}

\begin{remark}[Intuition on using pseudolabel]
Exploiting the pseudolabel serves the following intuition: Suppose that an input $x_i$ is misclassified. If there exists another input $x_i'$ which is close to $x_i$, and its pseudolabel $\hat y_i'$ is also close to $\hat y_i$, then it is also likely that $x_i'$ will also be misclassified. Thus, our construction of the VoI implicitly relies on the assumption that the pseudolabel is also informative to help predict failure patterns. If a user finds this assumption impractical, it is straightforward to remove this information from the specification of the kernel, and the construction of the Gaussian process still holds with minimal changes.
\end{remark}

\begin{remark}[Dimensionality reduction] If the feature space is high-dimensional ($d$ is large), we can apply a dimensionality reduction mapping $\varphi(x_i)$ to map the features to a space of smaller dimensions $d' \ll d$. The kernel $\mc K_{\mc Y}$ will be computed similarly with $\msa_y$ and $\covsa_y$ are all $d'$ dimensional.
\end{remark}


\subsection{Value Adjustment and Pattern Neighborhood Update} \label{sec:update}

At time $t$, the algorithm recommends querying the true label of the samples whose indices are in the set $\mathcal I_t$. If for $i \in \mc I_t$, the sample $x_i$ is misclassified, that is $\hat{y}_i \neq y_i^{\true}$, then we should update the observed probability to $\VoI(x_i, \hat{y}_i) = 1$. However, this would lead to updating $g(x_i, \hat{y}_i) = + \infty$. Similarly, if $x_i$ is classified correctly then we should update $g(x_i, \hat{y}_i) = - \infty$.
To alleviate the infinity issues, we set a finite bound $-\infty < L < 0 < U < +\infty$, and we update using the rule
\[
    g^{\text{observe}}(x_i, \hat{y}_i) = \begin{cases}
        U &\text{if $x_i$ is misclassified}, \\
        L &\text{otherwise.}
    \end{cases}
\]
Moreover, suppose that at time $t$, the annotator identifies that some samples form a pattern. In that case, we intentionally re-calibrate all the values for the samples in the pattern using
\[
    g^{\text{observe}}(x_i, \hat{y}_i) = L\quad \text{if $i$ belongs to a failure pattern}.
\]
In doing so, we intentionally adjust the misclassified sample $i$ to be a correctly classified (or `good') sample. 

\begin{remark}[Interpretation of VoI updates]
The VoI aims at capturing the probability that an \textit{un}labeled sample belongs to an \textit{un}identified failure mode. There are three exemplary cases to guide the design of the VoI: (i) if an unlabeled sample is near the misclassified samples and those misclassified samples do not form a failure mode yet, then VoI should be high. This signifies exploitation for confirmation: it is better to concentrate on the region surrounding this unlabeled sample to confirm this failure mode; (ii) if an unlabeled sample is near the correctly-classified samples, then VoI should be low. This is an exploration process: this sample may be in the correctly-classified region, and we should query in other regions;  (iii) if an unlabeled sample is near the misclassified samples and those misclassified samples already formed a failure mode, then VoI should be low. Again, this depicts an exploration process: it is better to search elsewhere for other failure modes.
\end{remark}


\section{Determinant Point Process Sampler for Annotation Recommendation} 
\label{sec:dpp}


Determinantal Point Processes (DDPs) is a family of stochastic models that originates from  quantum physics: DPPs are particularly useful to model the repulsive behavior of Fermion particles~\citep{ref:macchi1975coincidence}. Recently, DPPs have gained attraction in the machine learning community~\citep{ref:kulesza2012determinantal, affandi2014learning, urschel2017learning} thanks to its successes in recommendation systems~\citep{ref:chen2018fast, wilhelm2018practical, gartrell2017low} and text and video summarization~\citep{ref:lin2021learning, ref:cho2019multi, ref:gong2014diverse}, to name a few. Given $N$ samples, the corresponding DPP can be formalized as follows.

\begin{definition}[$L$-ensemble DPP] \label{def:dpp-L}
    Given a matrix $L \in \PSD^N$, an $L$-ensemble DPP is a distribution over all $2^N$ index subsets $J \subseteq \{1, \ldots, N\}$ such that
\[\mathrm{Prob}(J) = \det(L_J)/ \det(I + L),\]
where $L_J$ denotes the $|J|$-by-$|J|$ submatrix of $L$ with rows and columns indexed by $J$.
\end{definition}

In this paper, we construct a DPP to help select the next batch of samples to query the annotator for their true labels. We design the matrix $L$ that can balance between exploration (querying distant samples in order to identify \textit{new} potential failure patterns) and exploitation (querying neighborhood samples to confirm plausible failure patterns). Given $N$ samples, the exploration is determined by a similarity matrix $S \in \PSD^N$ that captures pairwise similarity of the samples
    \[
        S_{ij} = \kappa (x_i, x_j) \quad \forall (i, j)
    \]
    for some similarity metric $\kappa$. For example, we can use $\kappa \equiv \mathcal{K}_{\mathcal{X}}$, where $\mc K_{\mc X}$ is prescribed in~\eqref{eq:Kx} with the same bandwidth parameter $h_{\mc X}$.

    \textbf{Exploration.} At time $t$, conditioned on the samples that are already queried $\mathcal I_{[t]}$, we can recover a conditional similarity matrix $S_t^*$ for the set of \textit{un}queried samples following \citep[Proposition~1.2]{ref:borodin2005eynard}. Let $| \mc I_{[t]}|$ be the number of samples that have been drawn so far, then $S_t^*$ is a $(N - | \mc I_{[t]}|)$-dimensional positive (semi)definite matrix, calculated as
\[
    S_t^* = ([(S + I_{\mathcal{I}_{[t]}^*})^{-1}]_{\mathcal{I}_{[t]}^*})^{-1} - I.
\]
Thus, the matrix $S_t^*$ will serve as a diversity-promoting term of the conditional DPP at time $t$.

\textbf{Exploitation.} The exploitation is determined by a probability matrix $P \in \PSD^{N - | \mc I_{[t]}|}$. At time $t$, we can use $P_t^* = \mathrm{diag}(\gamma_{t,i})$, where $\gamma_{t,i}$ is the posterior probability of being misclassified defined in~\eqref{eq:mean}. Notice that if $\gamma_{t,i}$ is negative, we can replace $\gamma_{t,i}$ by the first-order approximation to guarantee that $P_t^*$ is a diagonal matrix with strictly positive diagonal elements. The matrix $P_t^*$ will induce exploitation because it promotes choosing samples with a high posterior probability of misclassification with the goal of confirming patterns.

\textbf{Exploration-Exploitation Balancing Conditional DPP.} We impose an additional parameter $\vartheta \in [0, 1]$ to capture the exploration-exploitation trade-off. At time $t$, we use a DPP with the kernel matrix $L^\vartheta$  defined as
    \[
        L_t^\vartheta = \vartheta S_t^* + (1-\vartheta) P_t^*
    \]
    for some mixture weight $\vartheta \in [0, 1]$. In particular, when $\vartheta$ is equal to zero, the algorithm's approach is entirely exploitative, and its primary objective is to confirm failure patterns in the dataset. Conversely, when $\vartheta$ is equal to one, the algorithm is entirely explorative, and its main aim is to recommend a diverse set of samples from the dataset. It is worth noting that the algorithm's behavior can be adjusted by modifying the value of $\vartheta$ to achieve an appropriate trade-off between exploration and exploitation depending on the specific problem at hand. Because both $S_t^*$ and $P_t^*$ are positive semidefinite matrices, the weighted matrix $L_t^\vartheta$ is also positive semidefinite, and specifies a valid DPP.

\textbf{Query Suggestion.}
We choose a set of unlabeled samples for annotation using a maximum a posteriori (MAP) estimate of the DPP specified by $L_t^\vartheta$.  
We then find the $s$ samples from the unlabeled data by solving the following problem
\be \label{eq:det}
    \max \left\{ \det (L_z) ~:~ z \in \{0, 1\}^{N - | \mc I_{[t]}|},~ \| z \|_0= s \right\},
\ee
where $L_z$ is a submatrix of $L_t^\vartheta \in \PSD^{N - | \mc I_{[t]}|}$ restricted to rows and columns indexed by the one-components of $z$. It is well-known that the solution to problem~\eqref{eq:det} coincides with the MAP estimate of the DPP with a cardinality constraint~\citep{ref:kulesza2012determinantal}. 

Unfortunately, problem~\eqref{eq:det} is NP-hard~\citep{ref:kulesza2012determinantal}. We thus use heuristics to find a good solution to the problem in a high-dimensional setting with a low running time. A common greedy algorithm to solve the MAP estimation problem is to iteratively find an index that maximizes the marginal gain to the incumbent set of chosen samples $z$. We then add the index $j$ to the set of samples until reaching the cardinality constraint of $s$ prototypes. This greedy construction algorithm has a complexity cost of $\mc O(s^2N)$ time for each inference. An implementation of this algorithm is provided in~\cite{ref:chen2018fast}. The greedy algorithm has been shown to achieve an approximation ratio of $\mc O(\frac{1}{s!})$\citep{ref:civril2009selecting}. Finally, to boost the solution quality, we add a $2$-neighborhood local search that swaps one element from the incumbent set with one element from the complementary set. This local search is performed until no further improvement is found.

\section{Bandwidth Selection} \label{sec:bandwidth} 

The product kernel $\mc K$ defined in~\eqref{eq:K} on the feature-pseudolabel label space requires the specification of two hyper-parameters: the bandwidth for the feature $h_{\mc X}$ and the bandwidth for the pseudolabels $h_{\mc Y}$. 
Given  $N$ feature-pseudolabel pairs $(x_i, \hat{y}_i)$ for $i = 1, \ldots, N$ and the kernel $\mc K$ defined as in Section~\ref{sec:GP}, we denote the Gram matrix by $K \in \PSD^N$ with the components of $K$ satisfying~\eqref{eq:Gram}. If the bandwidth parameters are too small compared to the feature distance $\| x - x'\|_2$ and the pseudolabel distance $\sqrt{ \| \msa_y - \msa_{y'} \|_2^2 + \| \covsa_y - \covsa_{y'} \|_F^2}$ in the dataset, then the matrix $K$ tends toward an $N$-by-$N$ identity matrix $I_N$. Notice that when $K$ is an identity matrix, the matrix multiplication $(K^*)^\top K^{-1}$ turns into a matrix of zeros, and the updates~\eqref{eq:update} become $m_t^* = m_{[t]}^*$ and $\cov_t^* = K_{[t]}^{**}$. This means that all observed information from previous queries is ignored. To alleviate this, we impose a restriction on $h_{\mc X}$ and $h_{\mc Y}$ so that
\be \label{eq:hyper-condition1}
    \| K(h_{\mc X}, h_{\mc Y}) - I_N \|_F \geq \delta \| I _N \|_F
\ee
for some value of $\delta > 0$. In the above equation, we make explicit the dependence of the Gram matrix $K$ on the hyper-parameters $h_{\mc X}$ and $h_{\mc Y}$, and the norm $\|\cdot \|_F$ is the Frobenius norm. Condition~\eqref{eq:hyper-condition1} imposes that the Gram matrix needs to be sufficiently different from the identity matrix, where the magnitude of the difference is controlled by $\delta$. The next proposition provides the condition to choose $h_{\mc X}$ and $h_{\mc Y}$ to satisfy this condition.

\begin{proposition}[Hyper-parameter values] \label{prop:hyper}
    For a fixed value of $\delta \in (0, \sqrt{N-1})$, the condition~\eqref{eq:hyper-condition1} is satisfied if
\[
\frac{D_{\mc X}}{h_{\mc X}^2} + \frac{D_{\mc Y}}{h_{\mc Y}^2} \leq \ln{\frac{N-1}{\delta^2}},
\]
where $\mc D_{\mc X}$ and $\mc D_{\mc Y}$ are calculated based on the dataset as
\begin{align*}
D_{\mc X} &= \frac{\sum_{i > j}\| x_i - x_j \|_2^2}{{N \choose 2}}, \quad \text{and} \\
D_{\mc Y} &= \frac{\sum_{i > j}\| \msa_{\hat{y}_i} - \msa_{\hat{y}_j} \|_2^2 + \| \covsa_{\hat{y}_i} - \covsa_{\hat{y}_j}\|_F^2}{{N \choose 2}}.
\end{align*}
\end{proposition}

We suggest choosing $h_{\mc X}$ and $h_{\mc X}$ to equalize the components
\[
\frac{D_{\mc X}}{h_{\mc X}^2} = \frac{D_{\mc Y}}{h_{\mc Y}^2} = \frac{1}{2} \ln{\frac{N-1}{\delta^2}}.
\]
We also notice that the value of $\delta = \sqrt{2} \times 10^{-6}$ is reasonable for most practical cases encountered in the numerical experiments of this paper. Hence, without other mention stated, we set $\delta$ to $\sqrt{2} \times 10^{-6}$.

\section{Numerical Experiments} 
\label{sec:experiment}

\textbf{Datasets.} For the numerical experiments, we utilize 15 real-world datasets adapted from~\citet{ref:eyuboglu2022domino}\footnote{Datasets are publicly available at~\url{https://dcbench.readthedocs.io/en/latest}}. Each dataset in \citet{ref:eyuboglu2022domino} consists of a pre-trained classifier and a collection of data points. Each data point has three features: Activation (a 512-dimensional embedding features of the image with respect to the pre-trained classifier), True Label (an integer in the range $\{0, \ldots, C\}$ that represents the true class of the data point), Probs (a $C$-dimensional vector that indicates the probability of each class). By taking the argmax of the Probs vector for each data point, we can determine the predicted label (pseudolabel) for that data point. 

We use the following construction of a failure pattern: Two samples are connected by an edge if each sample is in the $k_{\mathrm{nn}}$-nearest neighbors of the other. Because $\mc X$ is a feature space, we measure the distance between two samples by taking the Euclidean distance between $x_i$ and $x_j$. Notice that  $k_{\mathrm{nn}}$ is a parameter that is chosen by the user. Criterion C in Definition~\ref{def:pattern} is chosen as maximally connected subgraphs. Further discussion about this specific selection of the user is provided in Appendix~\ref{sec:specific-user}. Indicating the failure mode as above requires the user to input two hyper-parameters, $k_{nn}$ and $M$. The discussion about choosing values of $k_{nn}$ and $M$ in practical problems is in Appendix~\ref{sec:add_num_res}.

For each dataset, we construct a dataset that is suited for the task of failure identification as follows: we choose different values of $k_{nn}$ to construct the graph (cf.~Section~\ref{sec:problem}) and the evidence threshold $M$, then we generate the ground truth information about the true failure patterns by finding maximally connected components in $\mc G^{\mathrm{mis}}$. Each sample in the dataset is now augmented to have four features: Activation, True Label, Pseudo Label, and Pattern, where Pattern is an integer in the range $\{-1, 1, \ldots, P \}$, where $-1$ means that the sample does not belong to any failure pattern, and $P$ is the number of failure patterns in the dataset.

During the experiment, the true labels and patterns of samples are hidden: true labels are only accessible by querying the annotator, while pattern information is used to compute the performance ex-post. Our 15 generated datasets are classified into three classes based on the level of SNR: Low, Medium, and High. The details are in Appendix~\ref{sec:dataset}.

\textbf{Comparison.} We compare the following baselines:
\begin{itemize}[leftmargin=5mm]
    \item Active learning algorithms: We consider two popular active learning methods, namely BADGE~\citep{ash2019deep} and Coreset~\citep{sener2017active}. Because the classifier is fixed in our setting, the retraining stage in these active learning algorithms is omitted.
    \item Uniform Sampling (US) algorithm: At iteration $t$ with the set of remaining unlabeled samples $\mathcal{I}_{[t]}^*$, we pick a size $s$ subset of $\mathcal{I}_{[t]}^*$ with equal probability. This algorithm is a stochastic algorithm; hence we take results from 30 random seed numbers and calculate the average.
    \item Five variants of our Directed Sampling (DS\_$\vartheta$) algorithm, with $\vartheta$ chosen from $\{0, 0.25, 0.5, 0.75, 1\}$. At $\vartheta = 0$, our algorithm is purely exploitative, emphasizing the confirmation of failure patterns. At $\vartheta = 1$, our algorithm is purely exploration, emphasizing recommending diverse samples from the dataset.
\end{itemize} 

Throughout, the batch size is set to $s = 25$ for all competing methods. Codes and numerical results are available at~\url{https://github.com/nguyenngocbaocmt02/FPD}.

\textbf{Experiment 1 (Sensitivity).} The goal of this experiment is to measure the sensitivity of different recommendation algorithms. Hereby, sensitivity is defined as the fraction between the number of queried samples until the detection of the first failure pattern in the dataset and the total number of samples. This value measures how slowly we identify the first failure pattern: a lower sensitivity is more desirable.

We observed that the two active learning algorithms have the lowest performance. We suspect that the objective of active learning algorithms is to refine the decision boundaries to obtain better performance (accuracy), whereas the primary concern herein is to isolate misclassified clusters. Thus, active learning methods may not be applicable to the problem considered in this paper. 

We could see from Table~\ref{sensitivity} that the sensitivity of all methods decreases as the SNR increases. All DS variants except the extreme with $\vartheta = 1.0$ outperform the US method and active learning methods.  The poor performance of DS\_1.0 is attributed to the lack of an exploitative term, which is also the knowledge gathered from previous queries. Moreover, we notice that our proposed algorithm, DS\_0.25 and DS\_0 have the smallest sensitivity of 0.11 overall. While DS\_0.25 achieves the highest performance in datasets with low noise magnitude, DS\_0 is more effective in medium and high SNR contexts. This can be attributed to the fact that when the SNR is low, there are many noise samples. Consequently, if DS\_0 gets trapped in a noisy misclassified region, it may take considerable time to confirm whether this area contains any patterns because the algorithm is purely exploitative when $\vartheta = 0$. In contrast, DS\_0.25 overcomes this issue by incorporating an exploration term that avoids focusing too much on any area.

\begin{table}
\caption{Benchmark of Sensitivity on different noise magnitudes. Bolds indicate the best methods.}
\label{sensitivity}
{\fontsize{9}{13}\selectfont
\begin{tabular}{lllll}
\hline
         & \multicolumn{4}{c}{Noise Magnitude}                                               \\ \hline
Methods  & Low                & Medium             & High               & Overall            \\ \hline
US       & 0.48$\pm$0.13          & 0.29$\pm$0.07          & 0.22$\pm$0.07          & 0.33$\pm$0.15          \\ \hline
DS\_0    & 0.22$\pm$0.27          & \textbf{0.04$\pm$0.02} & \textbf{0.03$\pm$0.01} & \textbf{0.11$\pm$0.18}          \\ \hline
DS\_0.25 & \textbf{0.16$\pm$0.08} & 0.09$\pm$0.04          & 0.08$\pm$0.04          & \textbf{0.11$\pm$0.07} \\ \hline
DS\_0.5  & 0.27$\pm$0.20           & 0.15$\pm$0.16          & 0.07$\pm$0.04          & 0.16$\pm$0.17          \\ \hline
DS\_0.75 & 0.27$\pm$0.10           & 0.17$\pm$0.11          & 0.08$\pm$0.06          & 0.17$\pm$0.12          \\ \hline
DS\_1.0  & 0.50$\pm$0.10            & 0.35$\pm$0.12          & 0.22$\pm$0.07          & 0.36$\pm$0.15          \\ \hline
BADGE  & 0.56$\pm$0.10            & 0.35$\pm$0.07          & 0.27$\pm$0.06          & 0.40$\pm$0.15          \\ \hline
Coreset  & 0.62$\pm$0.08            & 0.44$\pm$0.07          & 0.27$\pm$0.08          & 0.44$\pm$0.16          \\ \hline
\end{tabular}
}
\end{table}

\textbf{Experiment 2 (Effectiveness).} This experiment aims to confirm the ability to recommend methods in detecting failure patterns subject to a limited number of annotation queries. More specifically, we allow the number of queries to be maximally $10\%$ and $20\%$ of the  total number of samples in the dataset, and we measure effectiveness as the percentage of detected patterns up to that cut-off. A higher value of effectiveness indicates a higher capacity for failure pattern identification.

When the maximum permitted number of queries is low (e.g., $10\%$), there is no significant difference in the overall performance of all algorithms because the queried information about the dataset is insufficient to confirm most patterns, see Table~\ref{0.1queries}. However, all versions of DS perform equally well and are more effective than the US, BADGE, and Coreset. As the number of queries increases to $20\%$ of the dataset in Table~\ref{0.2queries}, all DS variants significantly outperform US and active learning methods: the DS methods manage to detect more than a third of all failure patterns. In high SNR datasets, DS\_0.5 can even detect over half of the patterns on average.
\begin{table}[h]
\caption{Benchmark of Effectiveness (at $10\%$ of sample size) on different noise magnitudes. Bolds indicate the best methods.}
\label{0.1queries}
{\fontsize{9}{13}\selectfont
\begin{tabular}{lllll}
\hline
        & \multicolumn{4}{c}{Noise Magnitude}                                             \\ \hline
Methods & Low               & Medium             & High              & Overall            \\ \hline
US      & 0.00$\pm$0.00           & 0.00$\pm$0.01           & 0.02$\pm$0.08         & 0.01$\pm$0.05          \\ \hline
DS\_0.0  & \textbf{0.10$\pm$0.09} & \textbf{0.24$\pm$0.06} & \textbf{0.38$\pm$0.10} & \textbf{0.24$\pm$0.14} \\ \hline
DS\_0.25 & \textbf{0.10$\pm$0.13} & 0.18$\pm$0.19          & 0.22$\pm$0.19         & 0.17$\pm$0.18          \\ \hline
DS\_0.5  & 0.05$\pm$0.10          & 0.18$\pm$0.19          & 0.27$\pm$0.16         & 0.17$\pm$0.18          \\ \hline
DS\_0.75 & 0.00$\pm$0.00           & 0.10$\pm$0.12           & 0.32$\pm$0.19         & 0.14$\pm$0.18          \\ \hline
DS\_1.0  & 0.00$\pm$0.00           & 0.00$\pm$0.00            & 0.00$\pm$0.00           & 0.00$\pm$0.00            \\ \hline
BADGE  & 0.00$\pm$0.00           & 0.00$\pm$0.00            & 0.00$\pm$0.00           & 0.00$\pm$0.00            \\ \hline
Coreset  & 0.00$\pm$0.00           & 0.00$\pm$0.00            & 0.00$\pm$0.00           & 0.00$\pm$0.00            \\ \hline
\end{tabular}
}
\end{table}

\begin{table}
\caption{Benchmark of Effectiveness (at $20\%$ of sample size) on different noise magnitudes. Bolds indicate the best methods.}
\label{0.2queries}
{\fontsize{9}{13}\selectfont
\begin{tabular}{lllll}
\hline
        & \multicolumn{4}{c}{Noise Magnitude}                                               \\ \hline
Methods & Low                & Medium             & High               & Overall            \\ \hline
US      & 0.00$\pm$0.03           & 0.02$\pm$0.07          & 0.16$\pm$0.20           & 0.06$\pm$0.14          \\ \hline
DS\_0.0  & \textbf{0.29$\pm$0.21} & \textbf{0.31$\pm$0.18} & 0.38$\pm$0.1           & \textbf{0.33$\pm$0.18} \\ \hline
DS\_0.25 & 0.10$\pm$0.13           & 0.26$\pm$0.13          & 0.50$\pm$0.11           & 0.28$\pm$0.21          \\ \hline
DS\_0.5  & 0.14$\pm$0.13          & 0.27$\pm$0.25          & \textbf{0.52$\pm$0.15} & 0.31$\pm$0.24          \\ \hline
DS\_0.75 & 0.05$\pm$0.10           & 0.24$\pm$0.27          & 0.43$\pm$0.08          & 0.24$\pm$0.23          \\ \hline
DS\_1.0  & 0.00$\pm$0.00            & 0.05$\pm$0.10           & 0.15$\pm$0.20           & 0.07$\pm$0.14          \\ \hline
BADGE  & 0.00$\pm$0.00           & 0.01$\pm$0.05            & 0.06$\pm$0.14           & 0.02$\pm$0.09            \\ \hline
Coreset  & 0.00$\pm$0.00           & 0.00$\pm$0.00            & 0.05$\pm$0.10           & 0.02$\pm$0.06            \\ \hline
\end{tabular}
}
\end{table}

\textbf{Conclusions.} We proposed a sampling mechanism for the purpose of failure pattern identification. Given a classifier and a set of unlabeled data, the method sequentially suggests a batch of samples for annotation and then consolidates the information to detect the failure patterns. The sampling mechanism needs to balance two competing criteria: exploration (querying diverse samples to identify new potential failure patterns) and exploitation (querying neighborhood samples to collect evidence to confirm failure patterns). We constructed a Gaussian process to model the exploitative evolution of our belief about the failure patterns and used a DPP with a weighted matrix to balance the exploration-exploitation trade-off. The numerical experiments demonstrate that our sampling mechanisms outperform the uniform sampling method in both sensitivity and effectiveness measures.

\textbf{Acknowledgments. } We gratefully acknowledge the generous support from the CUHK's Improvement on Competitiveness in Hiring New Faculties Funding Scheme and the CUHK's Direct Grant Project Number 4055191.

\onecolumn 

\renewcommand\thefigure{\thesection.\arabic{figure}}    
\appendix
\section{Additional Experiments}
\label{Additional Experiments}
\subsection{Motivation for the specific user's defined failure mode definition}
\label{sec:specific-user}

\begin{figure}[h]
    \centering
    \includegraphics[scale=0.475]{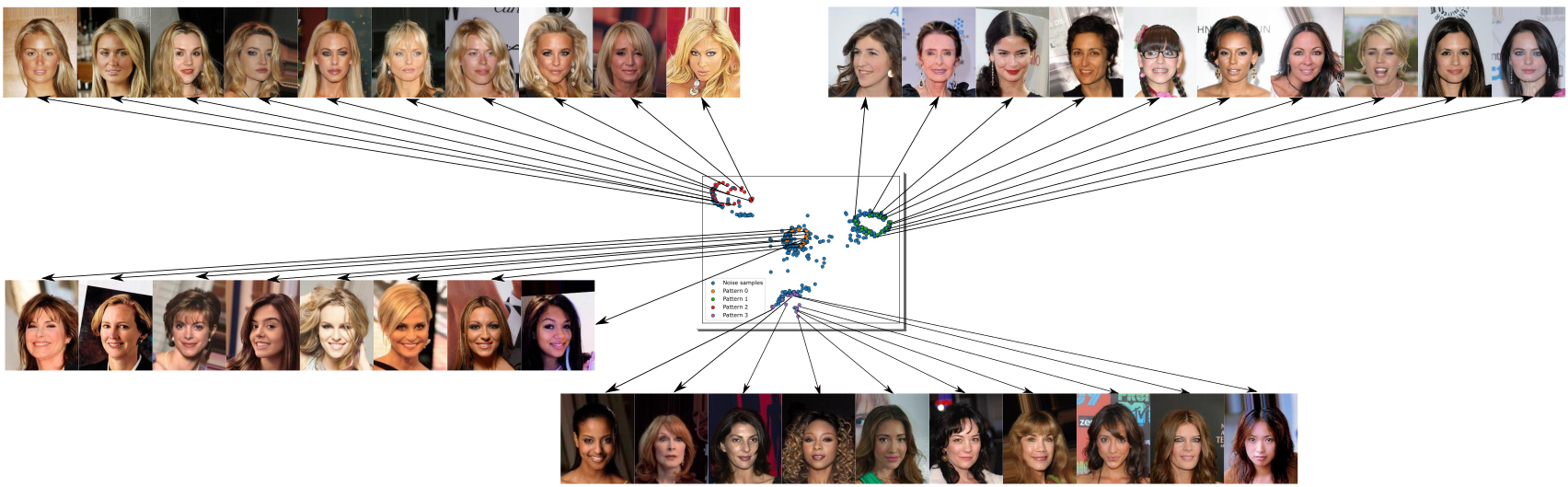}
    \caption{Failure patterns existing in dataset id\_1. One can observe four distinct failure patterns in this dataset.}
    \label{fig:gt}
\end{figure}

In this section, we provide the motivation, theoretical justification, and practical effectiveness of the failure mode definition based on mutual nearest neighbor graph\footnote{A mutual $k_{nn}$-nearest neighbor graph is a graph where there is an edge between $x_i$ and $x_j$ if  $x_i$ is one of the $k_{nn}$ nearest neighbors of  $x_j$ and $x_j$ is one of the $k_{nn}$ nearest neighbors of  $x_i$.} on embedding space.
\begin{itemize}
    \item We make a similar assumption with \citet{ref:d2022spotlight} and \citet{ref:sohoni2020no} that the classifier’s activations layer contains essential information about the semantic features used for classification. The proximity between two points in this embedding space could indicate their semantic similarity. Hence, issuing an edge between two points as in the mutual nearest neighbor graph likely guarantees that two connected points have much more semantic similarity than other pairs. This would ensure semantic cohesion for the points within a failure mode according to our definition.
    \item  Regarding the theoretical aspect, we use the mutual nearest neighbor graph, which is effective in clustering and outliers detection (see \citet{ref:song2022graph}, \citet{ref:song2022survey} and \citet{ref:brito1997connectivity}). Moreover, \citet[Theorem~2.2]{ref:brito1997connectivity} stated that with the reasonable choice of $k_{nn}$, connected components (a.k.a.~maximally connected subgraphs) in a mutual $k_{nn}$-graph are consistent for the identification of its clustering structure.
\item  In terms of more visual representation, we show images of four failure patterns of dataset id\_1 in Figure~\ref{fig:gt} to show the effectiveness of this definition on detecting semantic-cohesion clusters. We can observe that each failure pattern has a common concept recognizable by humans and includes images that are all misclassified. The top-left mode includes images of blonde-haired girls with tanned skin. The top-right mode includes images of girls wearing earrings. The bottom-left mode contains photos with tilted angles, and the bottom-right mode contains images with dark backgrounds.
\end{itemize}

\subsection{Datasets and Implementation Details}
\label{sec:dataset}

We describe fifteen datasets used in our work in Table~\ref{tab:dataset}. 

\begin{table}
\centering
\caption{The description of 15 datasets that are used in the numerical experiments.}
\begin{tabular}{cccccccc}
\hline
Dataset & DcBench   & Noise Magnitude       & SNR  & $M$  & $k_{nn}$ & Sample size & Number of misclassified samples \\ \hline
id\_1   & p\_72799  & Low     & 0.15 & 10 & 7     & 6088                  & 572                                \\ \hline
id\_2   & p\_122144 & Low     & 0.22 & 10 & 7     & 6103                  & 1076                               \\ \hline
id\_3   & p\_121880 & Low    & 0.38 & 10 & 7     & 5969                  & 1259                               \\ \hline
id\_4   & p\_122653 & Low    & 0.47 & 10 & 7     & 6019                  & 1088                               \\ \hline
id\_5   & p\_118660 & Low    & 0.47 & 10 & 8     & 5994                  & 1019                               \\ \hline
id\_6   & p\_122145 & Medium & 0.69 & 10 & 11    & 6135                  & 1141                               \\ \hline
id\_7   & p\_121753 & Medium & 0.96 & 10 & 10    & 6138                  & 1612                               \\ \hline
id\_8   & p\_122406 & Medium & 1.17 & 10 & 16    & 6072                  & 937                                \\ \hline
id\_9   & p\_118049 & Medium & 1.38 & 10 & 12    & 5979                  & 1051                               \\ \hline
id\_10  & p\_122150 & Medium & 1.39 & 10 & 10    & 6107                  & 1304                               \\ \hline
id\_11  & p\_121948 & High   & 1.75 & 10 & 15    & 6027                  & 1438                               \\ \hline
id\_12  & p\_122417 & High   & 1.85 & 10 & 19    & 6035                  & 1096                               \\ \hline
id\_13  & p\_122313 & High   & 1.91 & 10 & 15    & 6048                  & 1011                               \\ \hline
id\_14  & p\_121977 & High   & 2.19 & 10 & 17    & 6117                  & 1153                               \\ \hline
id\_15  & p\_121854 & High   & 3.78 & 10 & 24    & 6017                  & 1554                               \\ \hline
\end{tabular}
\label{tab:dataset}
\end{table}

\textbf{Preprocessing}: We single out 15 datasets from \cite{ref:eyuboglu2022domino}, each includes three features: Activation, True Label, and Pseudo Label. After that, we preprocess the data using a standard scaler for the Activation feature. 

\begin{figure}
	\begin{minipage}[t]{0.5\textwidth}
		\centering
		\includegraphics[scale = 0.63]{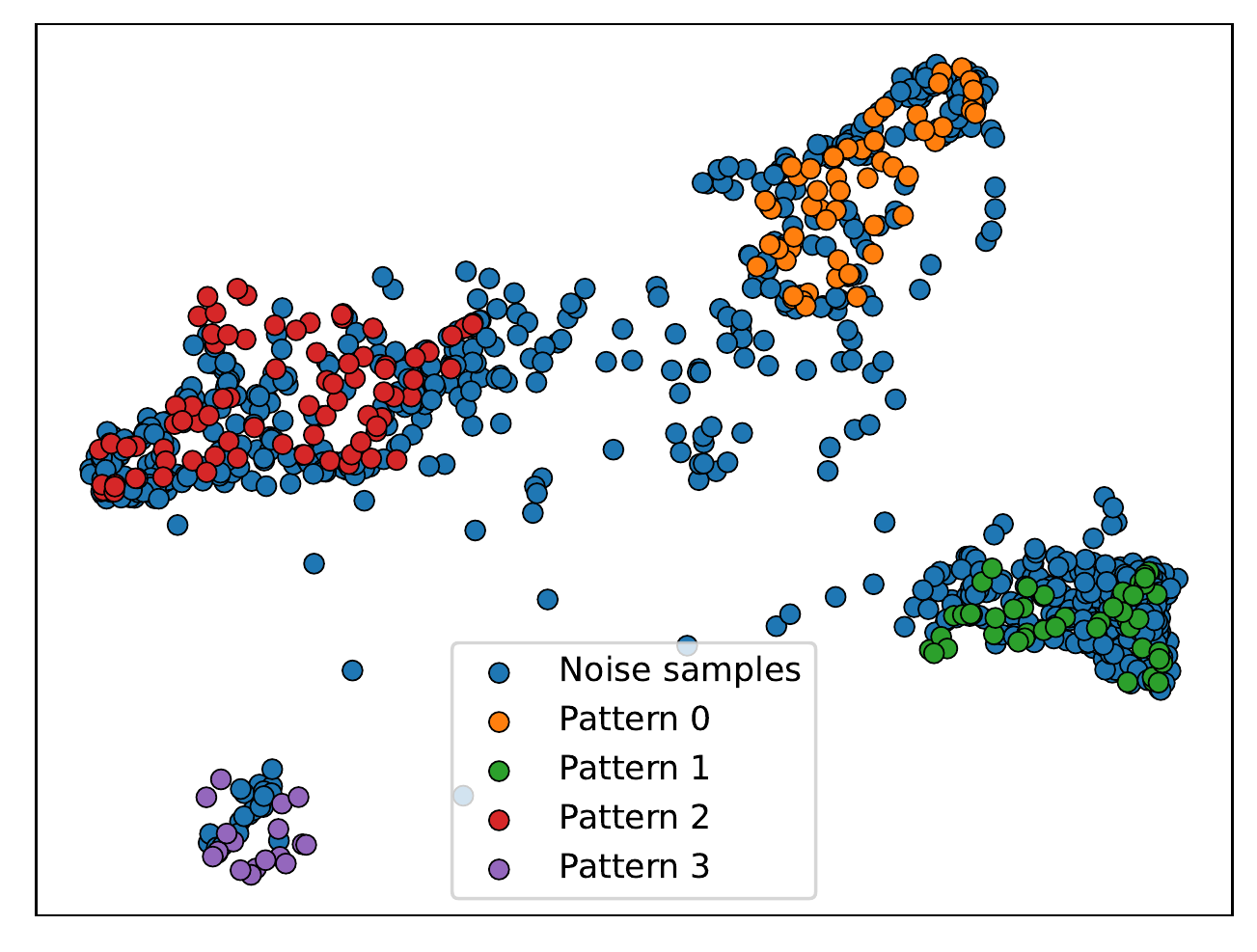}
        \caption*{id\_2 dataset (SNR = 0.22)}
	\end{minipage}
	\begin{minipage}[t]{0.5\textwidth}
		\centering
		\includegraphics[scale = 0.63]{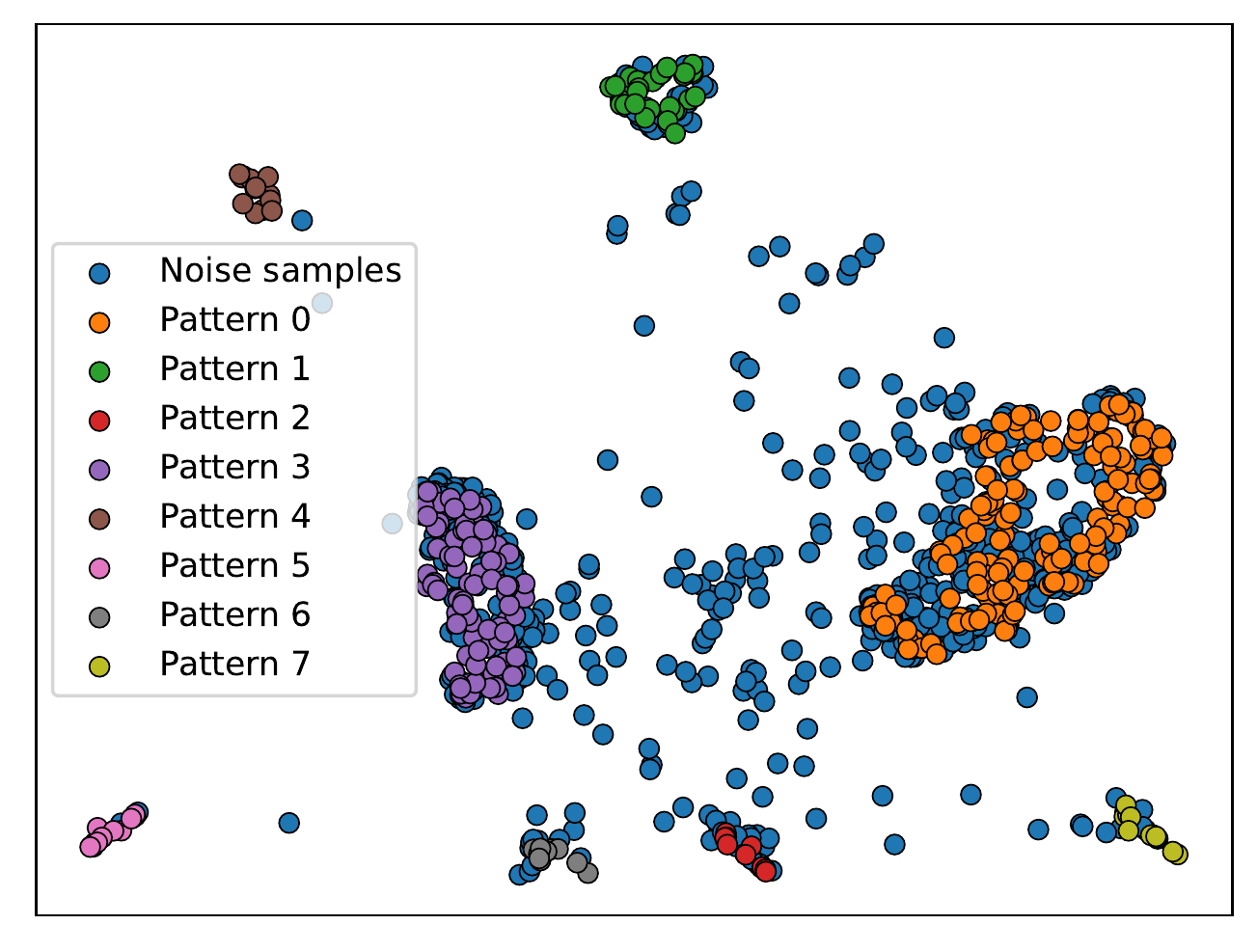}
        \caption*{id\_5 dataset (SNR = 0.47)}
	\end{minipage}
	\begin{minipage}[t]{0.5\textwidth}
		\centering
		\includegraphics[scale = 0.63]{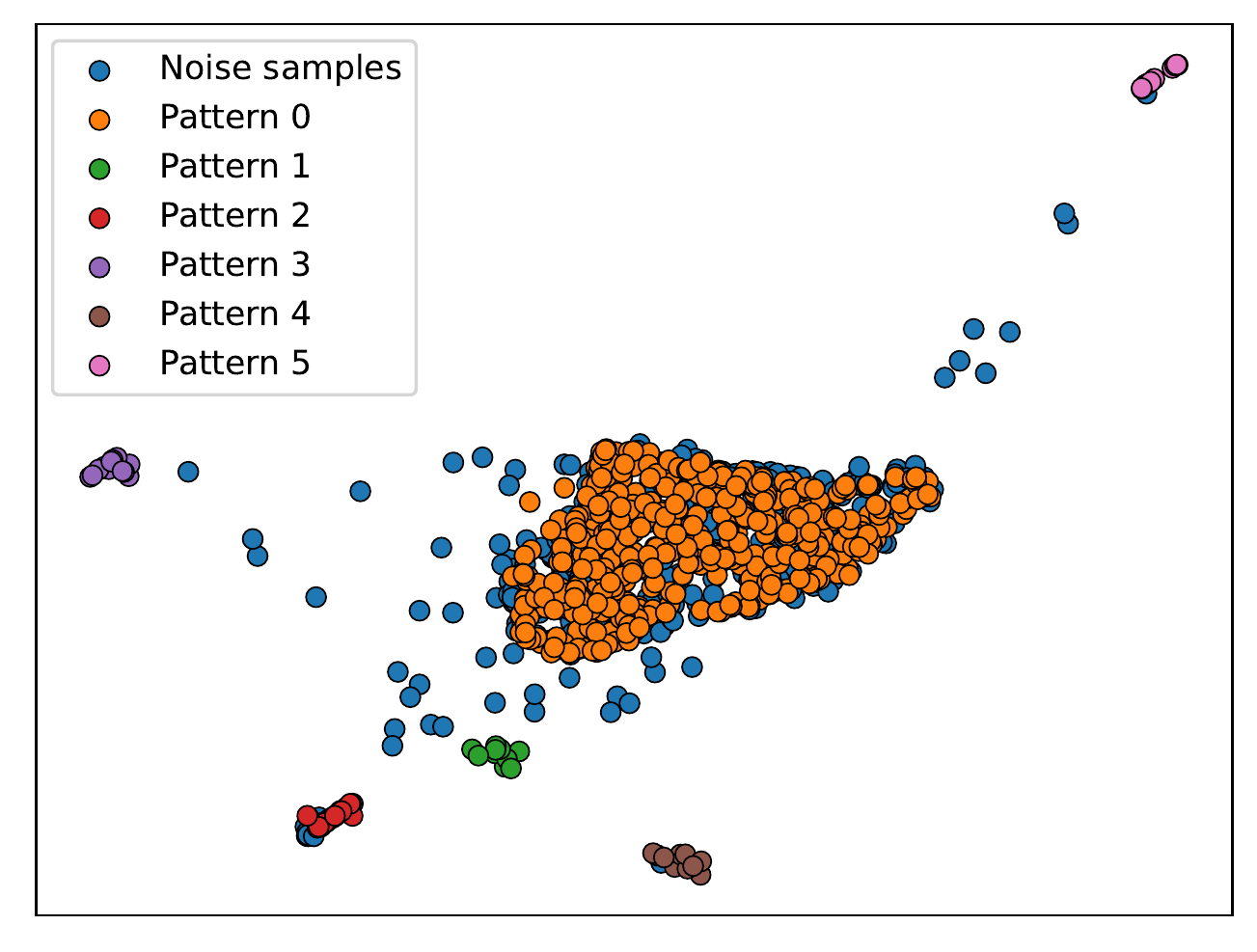}
        \caption*{id\_8 dataset (SNR = 1.17)}
	\end{minipage}
	\begin{minipage}[t]{0.5\textwidth}
		\centering
		\includegraphics[scale = 0.63]{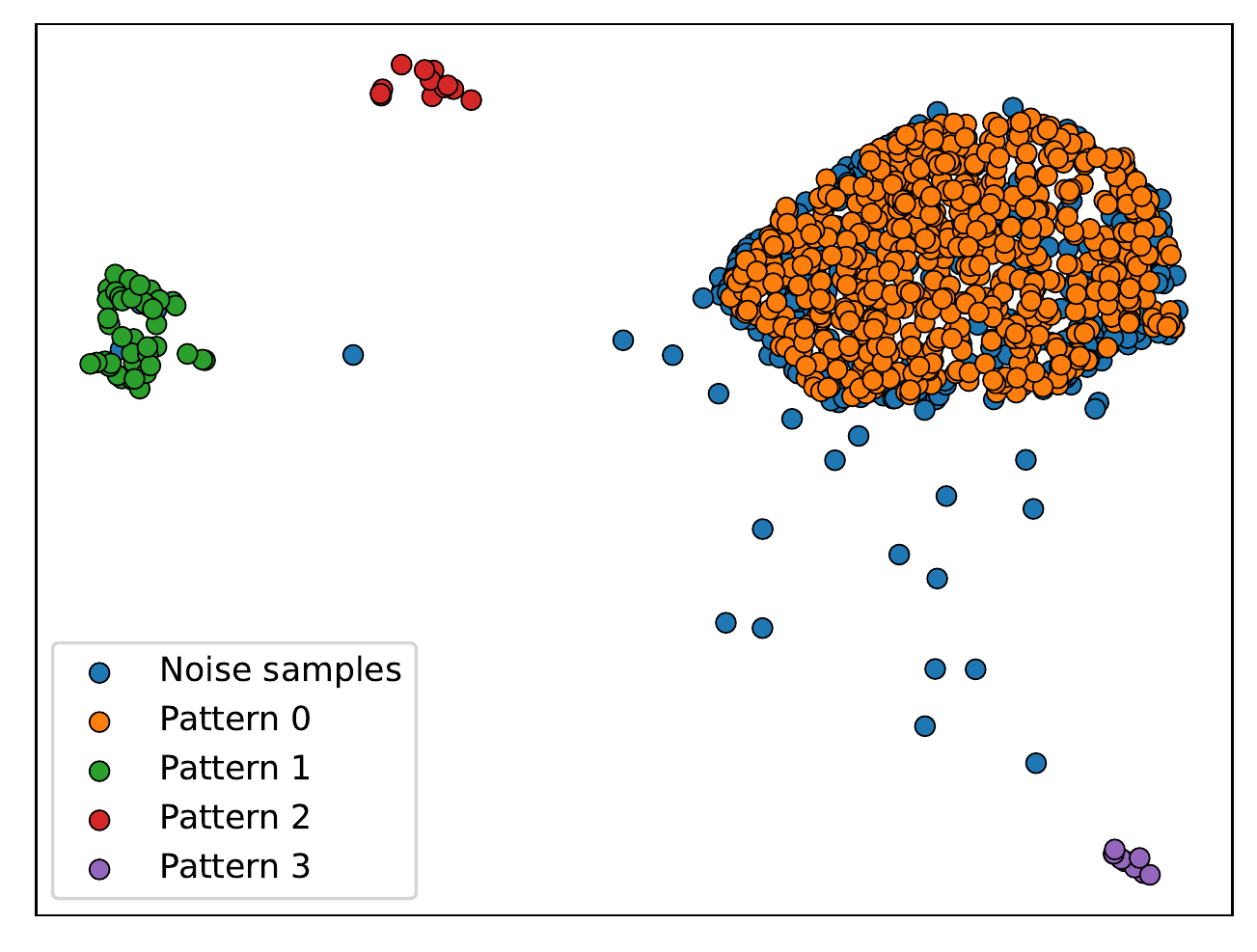}
        \caption*{id\_12 dataset (SNR = 1.85)}
	\end{minipage} 
   \caption{The 2-D visualization of the Activation feature in four datasets. To downsample from a 512-dimension vector to a 2-dimension vector, we utilize the Supervised Dimension Reduction technique introduced by \citet{ref:mcinnes2018umap}.}
    \label{fig:datasets}
\end{figure}

\textbf{Ground truth generation}: It is necessary to assign values of $k_{nn}$ and $M$ to each preprocessed dataset. The value of $M$ expresses the level of evidence required for confirming the failure patterns. A higher value of $M$ indicates a greater emphasis on the patterns that exist most frequently in the dataset. As $M$ decreases to 1, the problem transforms into identifying misclassified data points, where each failure data point constitutes a pattern. Moreover, the users choose $M$ so that they can perceive the shared concept of $M$ samples. If $M$ is too small, then the concept may not be distinctive enough between clusters, while if $M$ is too large, the users may have a bottleneck in identifying the shared concept. The value of $k_{nn}$ signifies the coherence required for data points within a pattern.  The users choose a smaller $k_{nn}$ if they need strong tightness between samples in a failure mode. \citet{ref:brito1997connectivity} recommended choosing $k_{nn}$ of order $\log(N)$ for consistent identification of the clustering structure. A smaller value of $k_{nn}$ imposes a more stringent condition to create an edge in the $k_{nn}$ graph. When $k_{nn}$ = 0, each data point is only connected with itself. If $k_{nn}$ is sufficiently high, all misclassified data points merge to form a single failure pattern. From Figure~\ref{fig:datasets}, we notice that as the increase of $k_{nn}$ and SNR, there is a tendency to appear big patterns with a large number of data points. We could explain it as follows. When increasing $k_{nn}$, more edges are additionally created, which could initially connect separate patterns or augment more data points into the patterns. In practical applications of this problem, it is important to note that the two parameters $k_{nn}$ and $M$ rely heavily on the users, the machine learning tasks, and the nature of the dataset. In this study, we have established a fixed value of M equal to 10 for all datasets, and we have varied the value of $k_{nn}$ to generate diverse scenarios of Signal-to-Noise Ratio (SNR). With the defined value of $k_{nn}$, we have constructed the $k_{nn}$ graph of the re-scaled Activation feature. Subsequently, we have employed a simple Depth First Search algorithm on the sub-graph of only misclassified data points to collect all maximally connected components with cardinality greater than $M$. These components represent patterns that are the focus of the recommending algorithms. We add one additional feature named Pattern to each data point which indicates the pattern of it or $-1$ if it does not belong to any patterns. 

Finally, the complete dataset for our problem consists of four information: Activation, True Label, Pseudo Label, and Pattern.

\subsection{Additional Numerical Results}
\label{sec:add_num_res}
In the main paper, we present the numerical results for groups categorized into three levels of Signal-to-Noise Ratio (SNR). In this section, we offer a comprehensive breakdown of the results for each individual dataset in Tables~\ref{tab:0.1detailed},~\ref{tab:0.2detailed}, and~\ref{tab:sensitivity}, respectively. 

\begin{figure}
	\begin{minipage}[t]{0.49\textwidth}
		\centering
		\includegraphics[scale = 0.43]{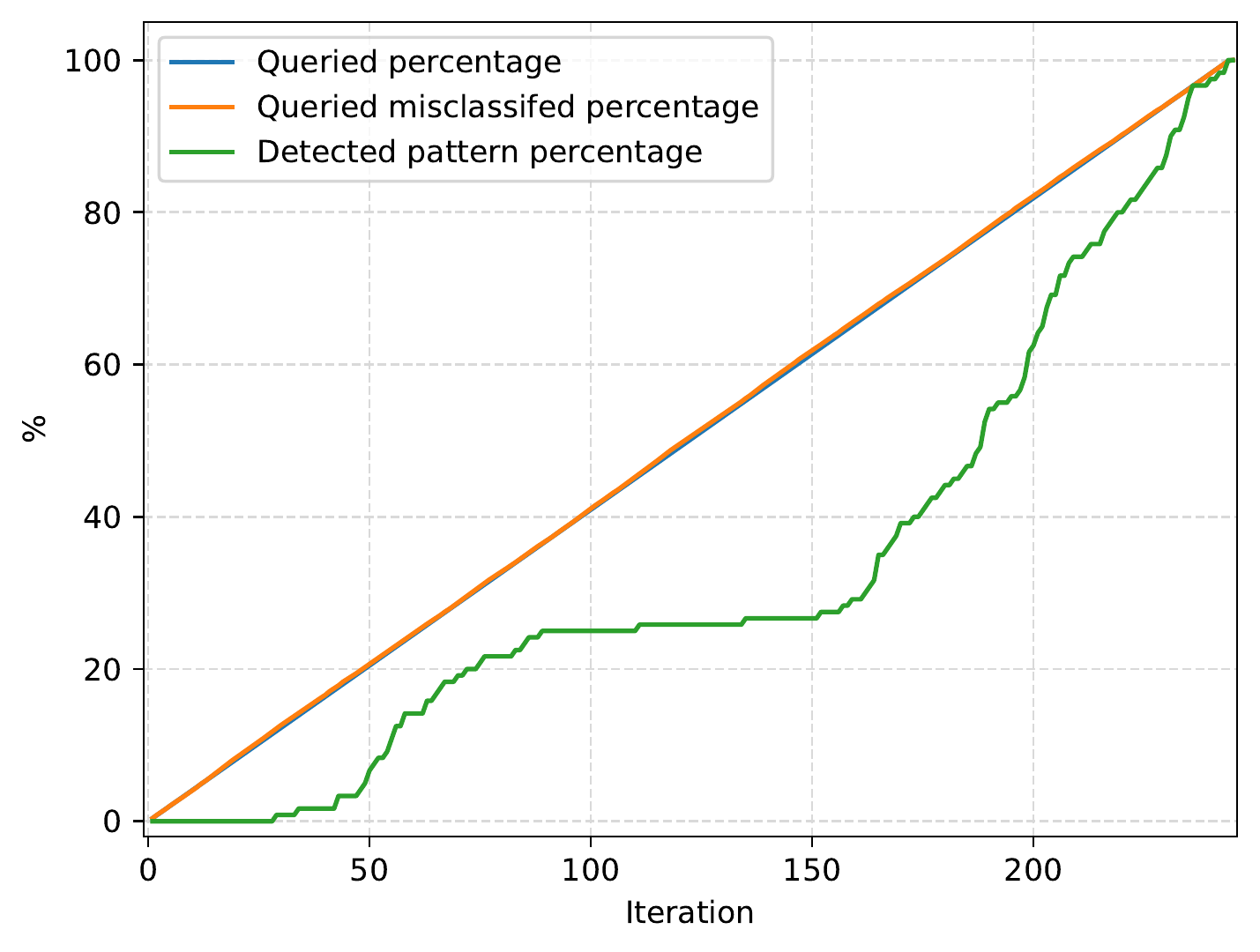}
            \caption*{US}
	\end{minipage}
	\hfill
	\begin{minipage}[t]{0.49\textwidth}
		\centering
		\includegraphics[scale = 0.43]{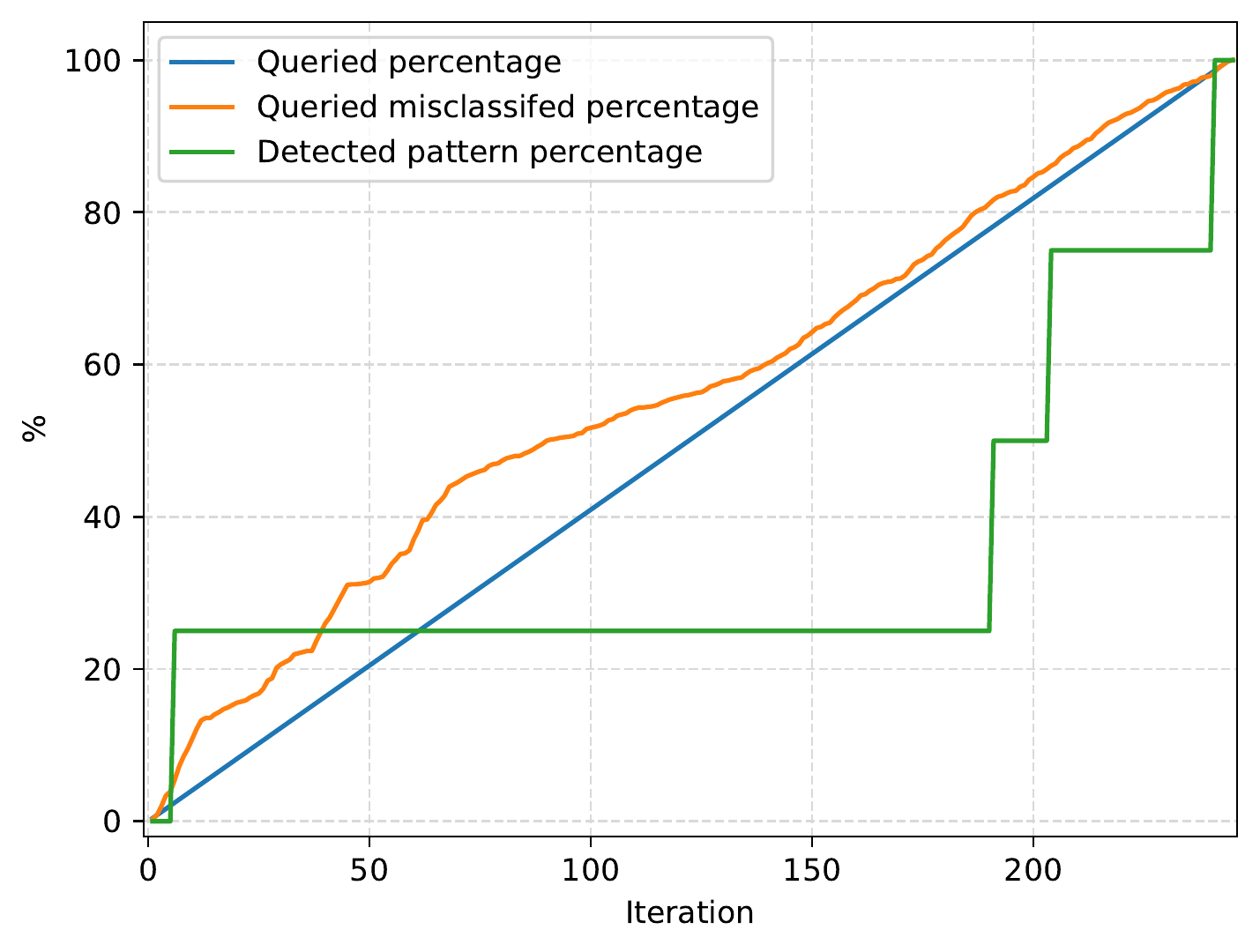}
            \caption*{DS\_0.0}
	\end{minipage}
	\hfill
	\begin{minipage}[t]{0.49\textwidth}
		\centering
		\includegraphics[scale = 0.43]{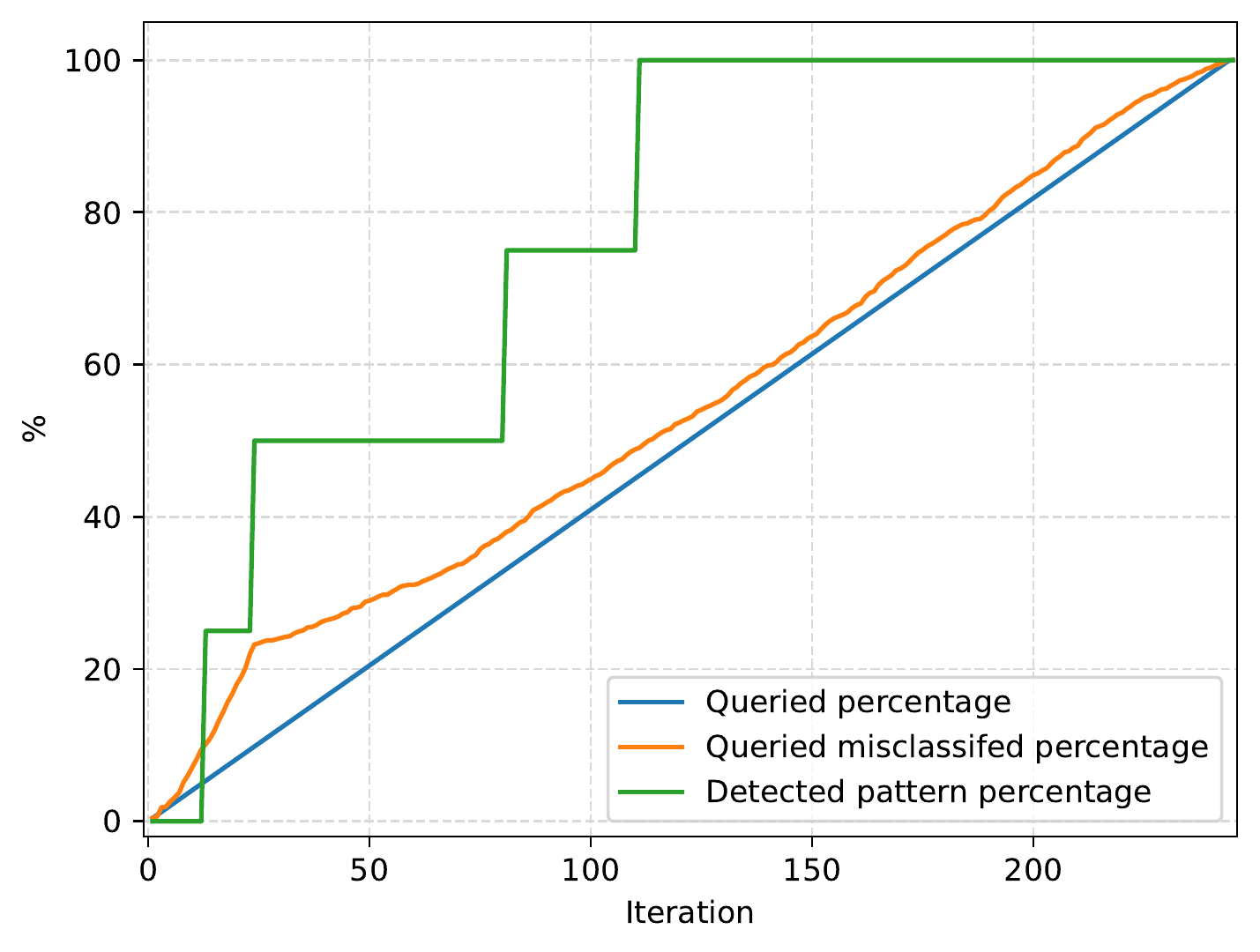}
            \caption*{DS\_0.25}
	\end{minipage}\hfill
	\begin{minipage}[t]{0.49\textwidth}
		\centering
		\includegraphics[scale = 0.43]{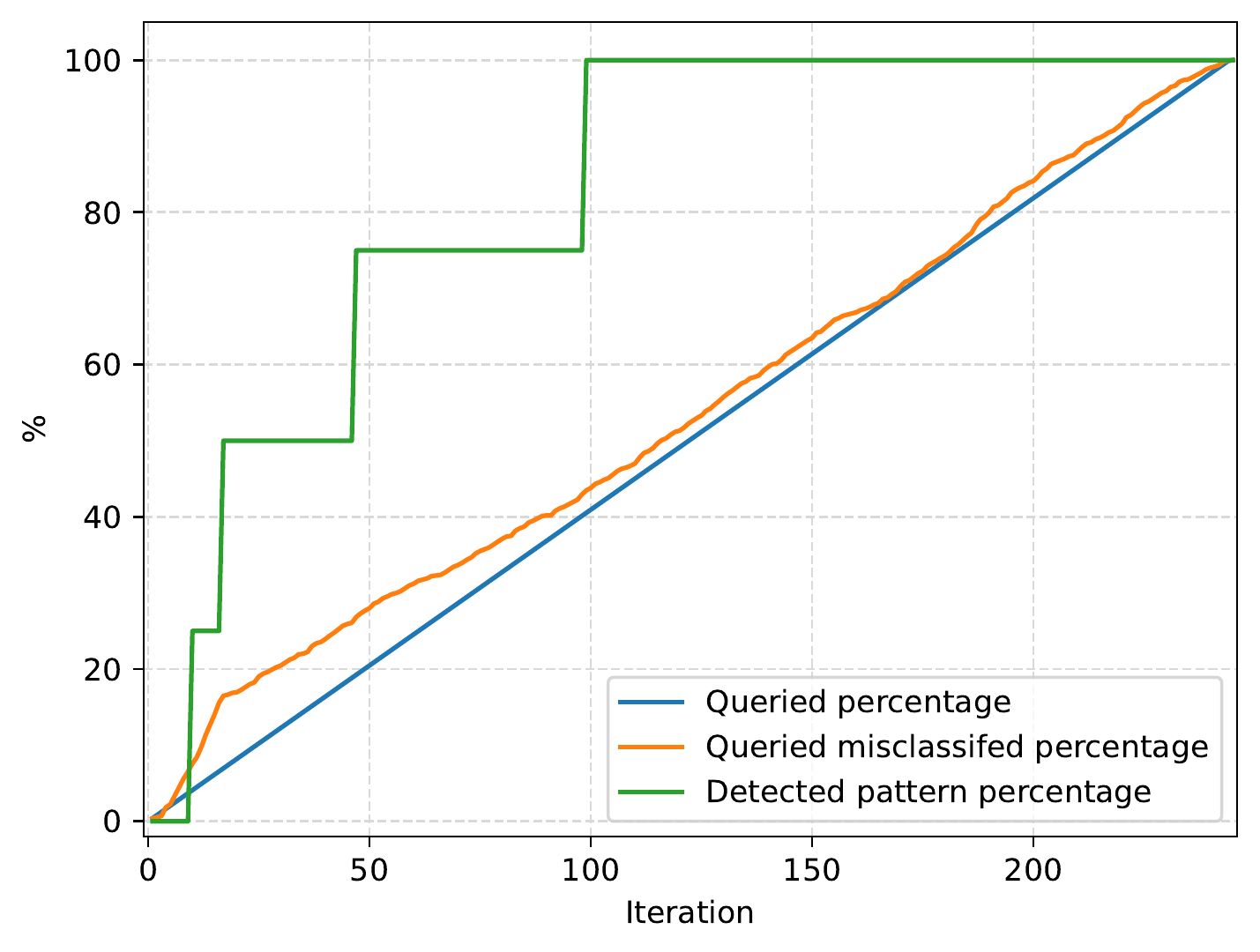}
            \caption*{DS\_0.5}
	\end{minipage}\hfill
	\begin{minipage}[t]{0.49\textwidth}
		\centering
		\includegraphics[scale = 0.43]{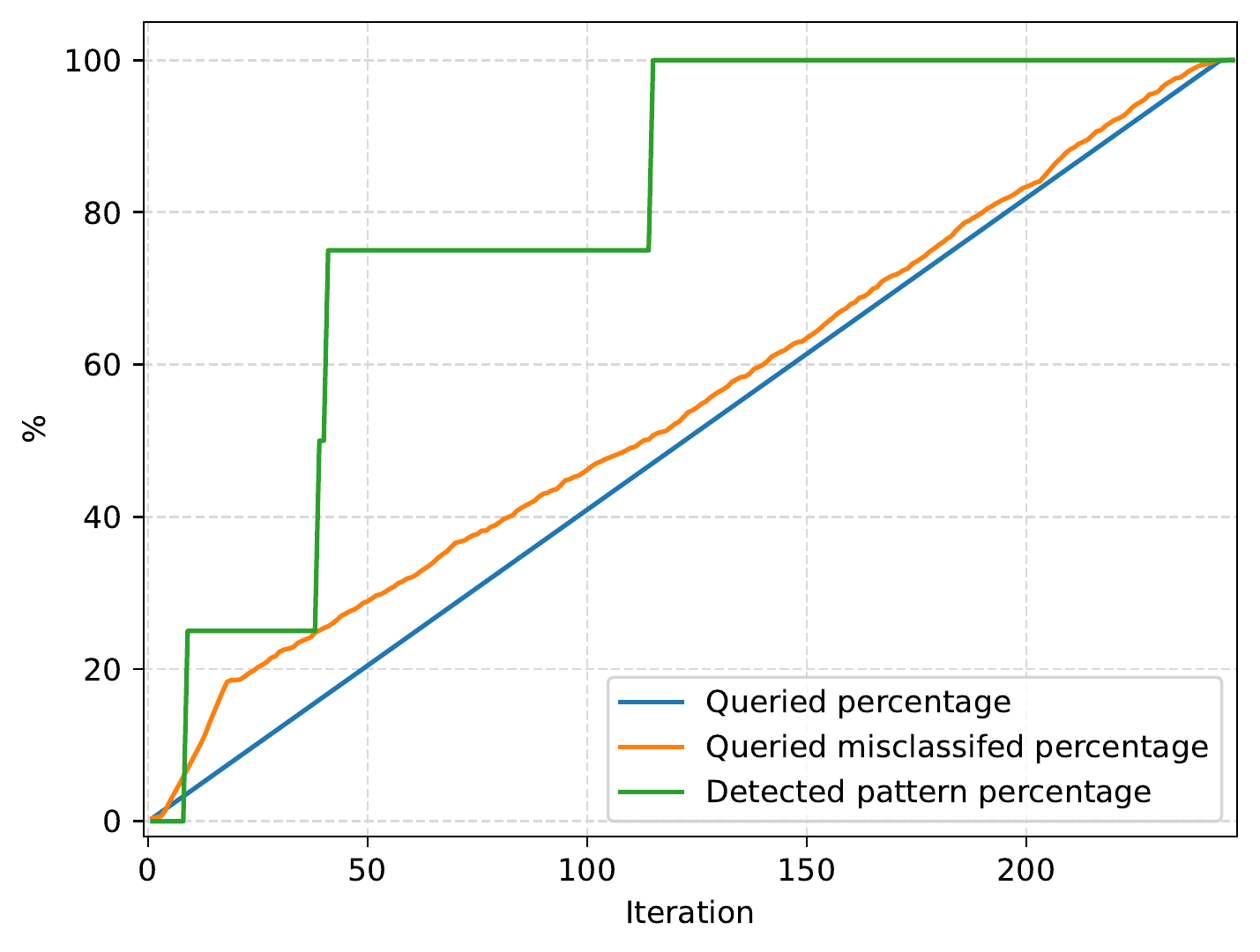}
            \caption*{DS\_0.75}
	\end{minipage}
    \hfill
	\begin{minipage}[t]{0.49\textwidth}
		\centering
		\includegraphics[scale = 0.43]{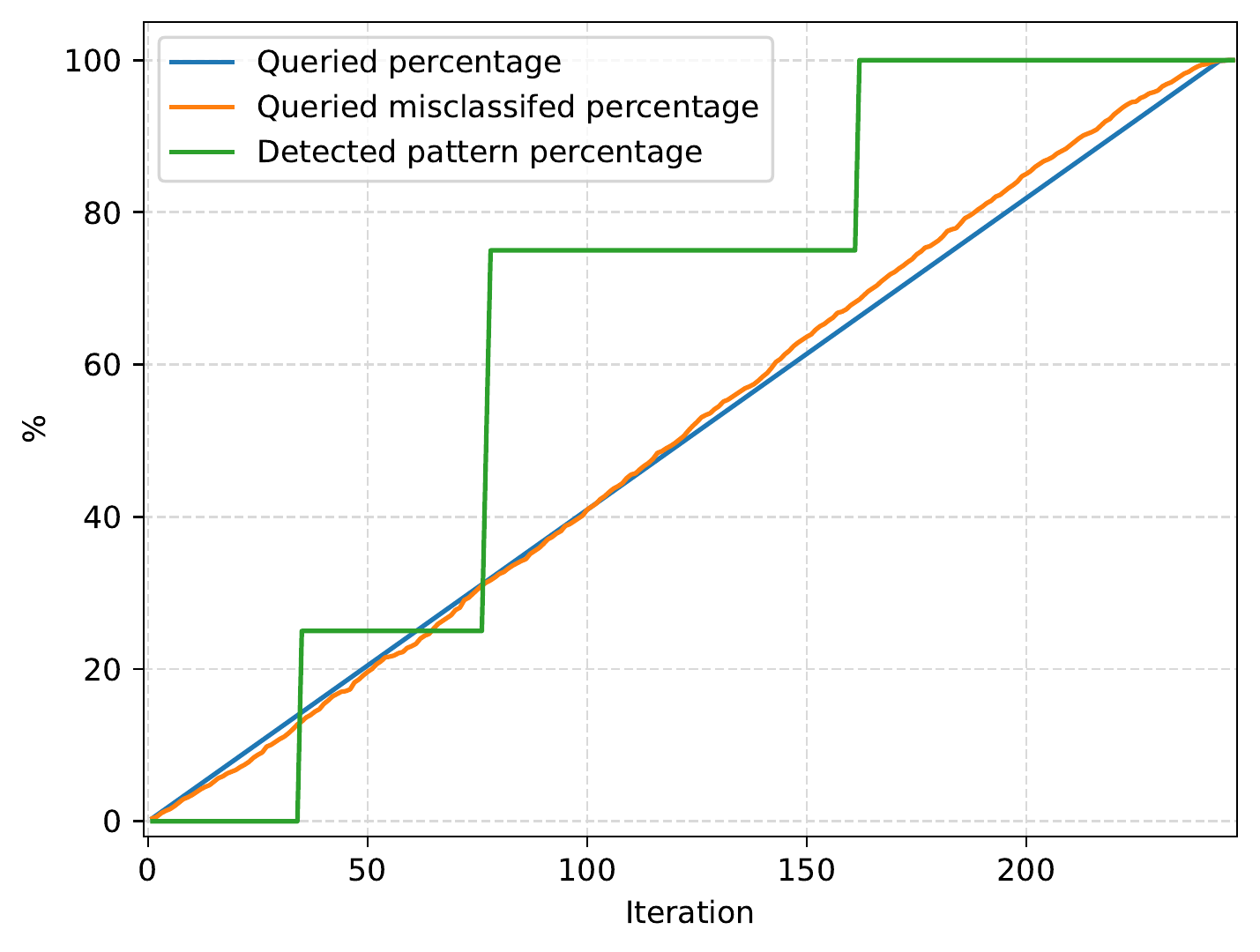}
            \caption*{DS\_1.0}
	\end{minipage}
	\hfill
 	\begin{minipage}[t]{0.49\textwidth}
		\centering
		\includegraphics[scale = 0.43]{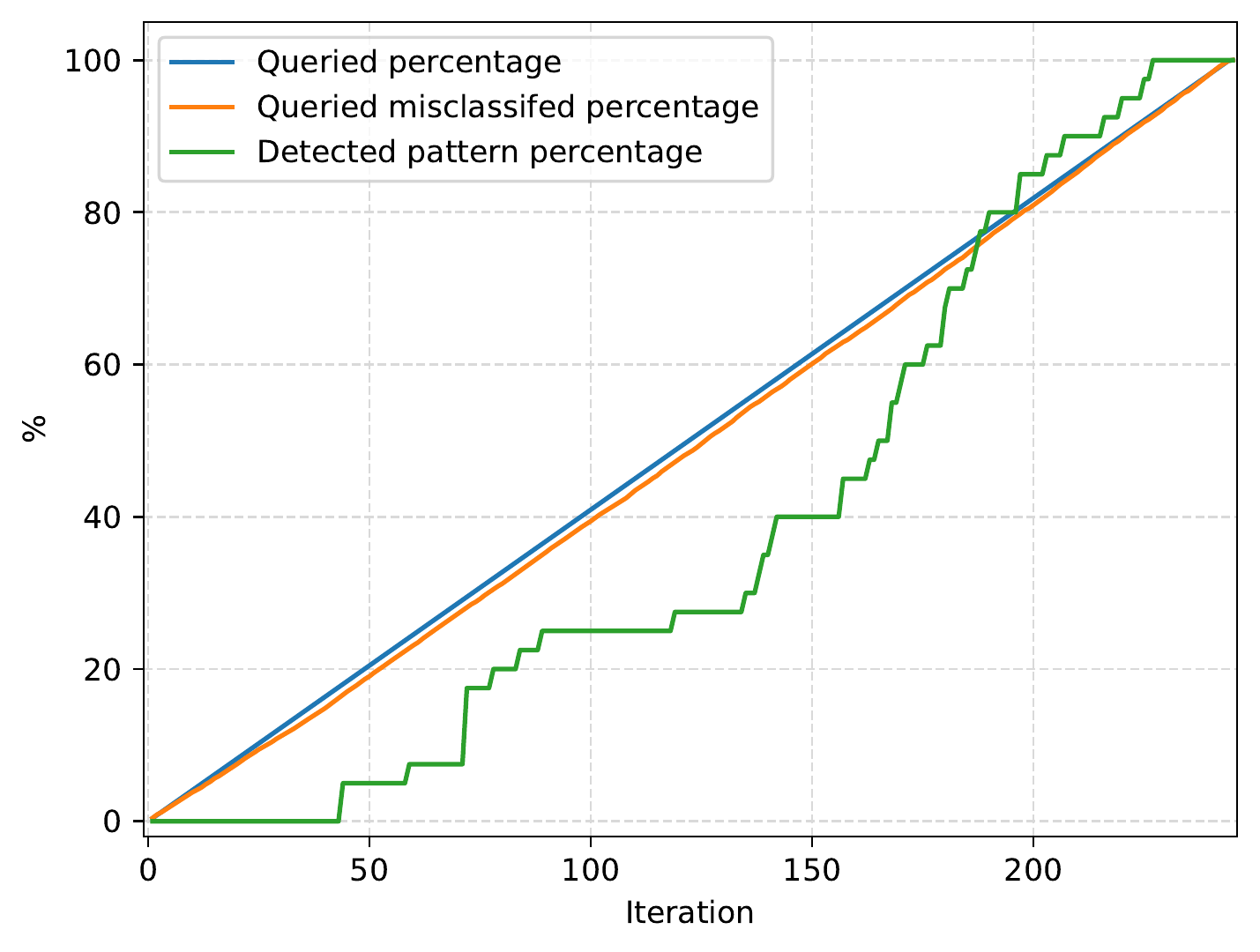}
            \caption*{BADGE}
	\end{minipage}
    \hfill
	\begin{minipage}[t]{0.49\textwidth}
		\centering
		\includegraphics[scale = 0.43]{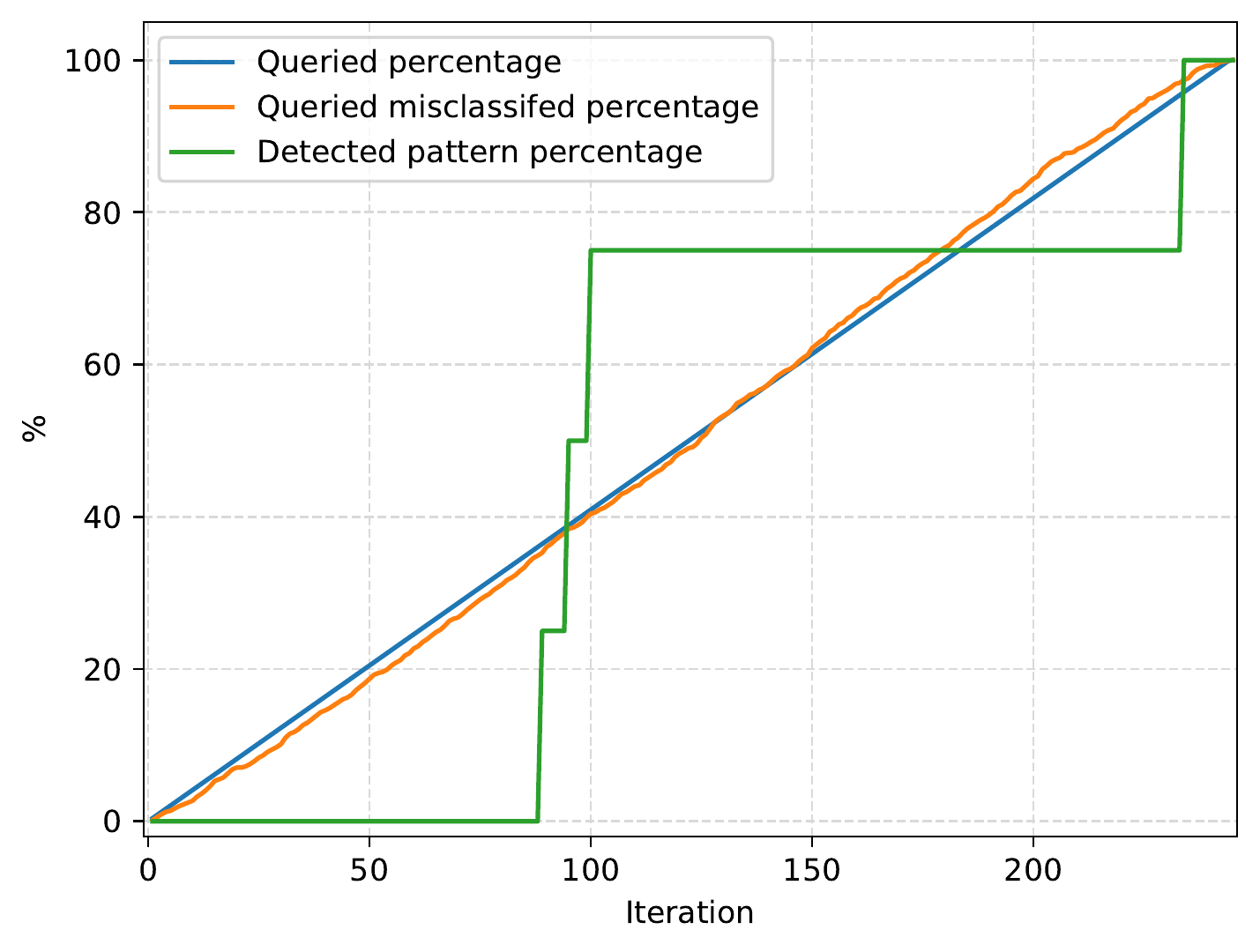}
            \caption*{Coreset}
	\end{minipage}
	\hfill
    \caption{The percentage of misclassified detected samples, the percentage of detected patterns, and the percentage of queried samples along with queried iterations in dataset id\_10}
    \label{fig:convergence}
\end{figure}

We also provide charts that illustrate the progress of algorithms over iterations in dataset id\_10, as depicted in Figure~\ref{fig:convergence}. The blue line represents the percentage of queried samples, which appears linear due to the fixed size of the queried batch at each iteration. The orange line indicates the percentage of detected misclassified samples out of the total misclassified ones in the dataset. The green line represents the percentage of detected failure modes out of the total number of failure modes in the dataset. It is evident that the orange line, corresponding to methods that incorporate our exploiting component (Gaussian process component) such as DS\_0.0, DS\_0.25, DS\_0.5, and DS\_0.75, consistently outperforms the blue lines significantly. This trend clearly demonstrates the effectiveness of our exploiting term in identifying misclassified samples.

However, DS\_0.0 shows inferior performance, as evidenced by the green line consistently falling below the blue line throughout the iterations, despite its effectiveness in identifying misclassified samples. In contrast, DS\_0.25, DS\_0.5, and DS\_0.75 exhibit superb performance in detecting all failure patterns within approximately 100 iterations (40\% of the dataset samples). This difference can be attributed to the absence of the exploration term in DS\_0.0 when dealing with a high SNR level in dataset id\_10.
\begin{table}
\centering
\caption{Benchmark of Effectiveness (at 10\% of sample size) on different noise magnitudes. Larger values are better.  Bolds indicate the best methods for each dataset.}
\begin{tabular}{ccccccccc}
\hline
Dataset & US & DS\_0.0 & DS\_0.25 & DS\_0.5 & DS\_0.75 & DS\_1.0 & Coreset & BADGE \\ \hline
id\_1 & 0.00±0.00 & 0.00±0.00 & 0.00±0.00 & 0.00±0.00 & 0.00±0.00 & 0.00±0.00 & 0.00±0.00 & 0.00±0.00 \\ \hline
id\_2 & 0.00±0.00 & \textbf{0.25±0.00} & 0.00±0.00 & 0.25±0.00 & 0.00±0.00 & 0.00±0.00 & 0.00±0.00 & 0.00±0.00 \\ \hline
id\_3 & 0.00±0.00 & \textbf{0.14±0.00} & \textbf{0.14±0.00} & 0.00±0.00 & 0.00±0.00 & 0.00±0.00 & 0.00±0.00 & 0.00±0.00 \\ \hline
id\_4 & 0.00±0.00 & 0.00±0.00 & \textbf{0.33±0.00} & 0.00±0.00 & 0.00±0.00 & 0.00±0.00 & 0.00±0.00 & 0.00±0.00 \\ \hline
id\_5 & 0.00±0.00 & \textbf{0.12±0.00} & 0.00±0.00 & 0.00±0.00 & 0.00±0.00 & 0.00±0.00 & 0.00±0.00 & 0.00±0.00 \\ \hline
id\_6 & 0.00±0.00 & \textbf{0.33±0.00} & 0.17±0.00 & 0.00±0.00 & 0.00±0.00 & 0.00±0.00 & 0.00±0.00 & 0.00±0.00 \\ \hline
id\_7 & 0.00±0.00 & \textbf{0.25±0.00} & \textbf{0.25±0.00} & \textbf{0.25±0.00} & \textbf{0.25±0.00} & 0.00±0.00 & 0.00±0.00 & 0.00±0.00 \\ \hline
id\_8 & 0.01±0.03 & \textbf{0.17±0.00} & 0.00±0.00 & \textbf{0.17±0.00} & 0.00±0.00 & 0.00±0.00 & 0.00±0.00 & 0.00±0.00 \\ \hline
id\_9 & 0.00±0.00 & \textbf{0.20±0.00} & 0.00±0.00 & 0.00±0.00 & 0.00±0.00 & 0.00±0.00 & 0.00±0.00 & 0.00±0.00 \\ \hline
id\_10 & 0.00±0.00 & 0.25±0.00 & \textbf{0.50±0.00} & \textbf{0.50±0.00} & 0.25±0.00 & 0.00±0.00 & 0.00±0.00 & 0.00±0.00 \\ \hline
id\_11 & 0.00±0.00 & \textbf{0.33±0.00} & 0.00±0.00 & 0.00±0.00 & \textbf{0.33±0.00} & 0.00±0.00 & 0.00±0.00 & 0.00±0.00 \\ \hline
id\_12 & 0.01±0.04 & \textbf{0.25±0.00} & \textbf{0.25±0.00} & \textbf{0.25±0.00} & \textbf{0.25±0.00} & 0.00±0.00 & 0.00±0.00 & 0.00±0.00 \\ \hline
id\_13 & 0.00±0.00 & \textbf{0.33±0.00} & \textbf{0.33±0.00} & \textbf{0.33±0.00} & 0.00±0.00 & 0.00±0.00 & 0.00±0.00 & 0.00±0.00 \\ \hline
id\_14 & 0.01±0.04 & \textbf{0.50±0.00} & 0.00±0.00 & 0.25±0.00 & \textbf{0.50±0.00} & 0.00±0.00 & 0.00±0.00 & 0.00±0.00 \\ \hline
id\_15 & 0.07±0.17 & \textbf{0.50±0.00} & \textbf{0.50±0.00} & \textbf{0.50±0.00} & \textbf{0.50±0.00} & 0.00±0.00 & 0.00±0.00 & 0.00±0.00 \\ \hline
Overall & 0.01±0.05 & \textbf{0.24±0.14} & 0.17±0.18 & 0.17±0.18 & 0.14±0.18 & 0.00±0.00 & 0.00±0.00 & 0.00±0.00 \\ \hline
\end{tabular}
\label{tab:0.1detailed}
\end{table}

\begin{table}
\caption{Benchmark of Effectiveness (at 20\% of sample size) on different noise magnitudes. Larger values are better.  Bolds indicate the best methods for each dataset.}
\centering
\begin{tabular}{ccccccccc}
\hline
Dataset & US & DS\_0.0 & DS\_0.25 & DS\_0.5 & DS\_0.75 & DS\_1.0 & Coreset & BADGE \\ \hline
id\_1 & 0.00±0.00 & 0.00±0.00 & 0.00±0.00 & 0.00±0.00 & 0.00±0.00 & 0.00±0.00 & 0.00±0.00 & 0.000±0.000 \\ \hline
id\_2 & 0.01±0.04 & \textbf{0.25±0.00} & 0.00±0.00 & \textbf{0.25±0.00} & \textbf{0.25±0.00} & 0.00±0.00 & 0.00±0.00 & 0.000±0.000 \\ \hline
id\_3 & 0.00±0.00 & \textbf{0.29±0.00} & 0.14±0.00 & 0.00±0.00 & 0.00±0.00 & 0.00±0.00 & 0.00±0.00 & 0.000±0.000 \\ \hline
id\_4 & 0.01±0.06 & \textbf{0.67±0.00} & 0.33±0.00 & 0.33±0.00 & 0.00±0.00 & 0.00±0.00 & 0.00±0.00 & 0.000±0.000 \\ \hline
id\_5 & 0.00±0.00 & \textbf{0.25±0.00} & 0.00±0.00 & 0.12±0.00 & 0.00±0.00 & 0.00±0.00 & 0.00±0.00 & 0.000±0.000 \\ \hline
id\_6 & 0.00±0.00 & \textbf{0.67±0.00} & 0.17±0.00 & 0.00±0.00 & 0.00±0.00 & 0.00±0.00 & 0.00±0.00 & 0.000±0.000 \\ \hline
id\_7 & 0.02±0.08 & \textbf{0.25±0.00} & \textbf{0.25±0.00} & \textbf{0.25±0.00} & \textbf{0.25±0.00} & 0.00±0.00 & 0.00±0.00 & 0.000±0.000 \\ \hline
id\_8 & 0.01±0.03 & \textbf{0.17±0.00} & \textbf{0.17±0.00} & \textbf{0.17±0.00} & 0.00±0.00 & 0.00±0.00 & 0.00±0.00 & 0.000±0.000 \\ \hline
id\_9 & 0.02±0.06 & \textbf{0.20±0.00} & \textbf{0.20±0.00} & \textbf{0.20±0.00} & \textbf{0.20±0.00} & 0.00±0.00 & 0.00±0.00 & 0.000±0.000 \\ \hline
id\_10 & 0.05±0.10 & 0.25±0.00 & 0.50±0.00 & \textbf{0.75±0.00} & \textbf{0.75±0.00} & 0.25±0.00 & 0.00±0.00 & 0.005±0.100 \\ \hline
id\_11 & 0.06±0.12 & 0.33±0.00 & 0.33±0.00 & \textbf{0.67±0.00} & 0.33±0.00 & 0.00±0.00 & 0.00±0.00 & 0.003±0.100 \\ \hline
id\_12 & 0.12±0.12 & 0.25±0.00 & \textbf{0.50±0.00} & 0.25±0.00 & \textbf{0.50±0.00} & 0.00±0.00 & 0.00±0.00 & 0.000±0.000 \\ \hline
id\_13 & 0.04±0.11 & 0.33±0.00 & \textbf{0.67±0.00} & \textbf{0.67±0.00} & 0.33±0.00 & 0.00±0.00 & 0.00±0.00 & 0.000±0.000 \\ \hline
id\_14 & 0.11±0.12 & \textbf{0.50±0.00} & \textbf{0.50±0.00} & \textbf{0.50±0.00} & \textbf{0.50±0.00} & 0.25±0.00 & 0.25±0.00 & 0.100±0.120 \\ \hline
id\_15 & 0.48±0.09 & \textbf{0.50±0.00} & \textbf{0.50±0.00} & \textbf{0.50±0.00} & \textbf{0.50±0.00} & 0.50±0.00 & 0.00±0.00 & 0.150±0.230 \\ \hline
Overall & 0.06±0.14 & \textbf{0.33±0.18} & 0.28±0.21 & 0.31±0.24 & 0.24±0.23 & 0.07±0.14 & 0.02±0.06 & 0.020±0.090 \\ \hline
\end{tabular}
\label{tab:0.2detailed}
\end{table}

\begin{table}
\caption{Benchmark of Sensitivity on different noise magnitudes. Smaller values are better. Bolds indicate the best methods in each dataset.}
\centering
\begin{tabular}{ccccccccc}
\hline
Dataset & US & DS\_0.0 & DS\_0.25 & DS\_0.5 & DS\_0.75 & DS\_1.0 & Coreset & BADGE \\ \hline
id\_1 & 0.55±0.00 & 0.76±0.00 & \textbf{0.23±0.00} & 0.57±0.00 & 0.44±0.00 & 0.62±0.00 & 0.76±0.00 & 0.660±0.009 \\ \hline
id\_2 & 0.35±0.00 & 0.09±0.00 & 0.23±0.00 & \textbf{0.05±0.00} & 0.14±0.00 & 0.51±0.00 & 0.60±0.00 & 0.600±0.100 \\ \hline
id\_3 & 0.59±0.00 & \textbf{0.05±0.00} & 0.07±0.00 & 0.42±0.00 & 0.31±0.00 & 0.49±0.00 & 0.55±0.00 & 0.550±0.007 \\ \hline
id\_4 & 0.44±0.00 & 0.13±0.00 & \textbf{0.06±0.00} & 0.11±0.00 & 0.22±0.00 & 0.57±0.00 & 0.60±0.00 & 0.530±0.006 \\ \hline
id\_5 & 0.53±0.00 & \textbf{0.08±0.00} & 0.21±0.00 & 0.19±0.00 & 0.23±0.00 & 0.33±0.00 & 0.57±0.00 & 0.470±0.006 \\ \hline
id\_6 & 0.48±0.00 & \textbf{0.03±0.00} & 0.04±0.00 & 0.47±0.00 & 0.29±0.00 & 0.32±0.00 & 0.55±0.00 & 0.400±0.007 \\ \hline
id\_7 & 0.37±0.00 & \textbf{0.03±0.00} & 0.10±0.00 & 0.07±0.00 & 0.07±0.00 & 0.35±0.00 & 0.37±0.00 & 0.340±0.004 \\ \hline
id\_8 & 0.30±0.00 & 0.08±0.00 & 0.12±0.00 & \textbf{0.03±0.00} & 0.30±0.00 & 0.46±0.00 & 0.45±0.00 & 0.400±0.006 \\ \hline
id\_9 & 0.28±0.00 & \textbf{0.03±0.00} & 0.14±0.00 & 0.13±0.00 & 0.15±0.00 & 0.45±0.00 & 0.48±0.00 & 0.330±0.005 \\ \hline
id\_10 & 0.20±0.00 & \textbf{0.02±0.00} & 0.05±0.00 & 0.04±0.00 & 0.04±0.00 & 0.14±0.00 & 0.36±0.00 & 0.280±0.006 \\ \hline
id\_11 & 0.26±0.00 & \textbf{0.01±0.00} & 0.13±0.00 & 0.15±0.00 & 0.06±0.00 & 0.31±0.00 & 0.32±0.00 & 0.300±0.005 \\ \hline
id\_12 & 0.13±0.00 & \textbf{0.02±0.00} & 0.04±0.00 & 0.04±0.00 & 0.07±0.00 & 0.21±0.00 & 0.22±0.00 & 0.290±0.003 \\ \hline
id\_13 & 0.22±0.00 & 0.05±0.00 & 0.05±0.00 & \textbf{0.04±0.00} & 0.19±0.00 & 0.29±0.00 & 0.40±0.00 & 0.340±0.006 \\ \hline
id\_14 & 0.18±0.00 & \textbf{0.03±0.00} & 0.11±0.00 & 0.10±0.00 & \textbf{0.03±0.00} & 0.11±0.00 & 0.18±0.00 & 0.230±0.004 \\ \hline
id\_15 & 0.23±0.00 & \textbf{0.02±0.00} & 0.05±0.00 & 0.05±0.00 & 0.04±0.00 & 0.19±0.00 & 0.26±0.00 & 0.210±0.003 \\ \hline
Overall & 0.34±0.14 & \textbf{0.10±0.18} & 0.11±0.07 & 0.16±0.17 & 0.17±0.12 & 0.36±0.15 & 0.44±0.16 & 0.400±0.150 \\ \hline
\end{tabular}
\label{tab:sensitivity}
\end{table}

\subsection{Analysis of sampling complexity}
Each iteration in our framework consists of two main phases. The first phase determines which samples to be labeled next, the most costly computation in this phase is the matrix inversion and computing matrix determinant. The maximum size of the matrix is $N$, so the time complexity is $O(N^3)$. If we use the optimized CW-like algorithm for matrix inversion, then the complexity can be as low as $O(N^{2.373})$. The second phase includes updating information and confirming detected failure modes. Updating information involves matrix inversions and multiplications, with cost $O(N^{2.373})$. A low-cost Depth First Search is implemented to check detected failure modes, which costs $O(N)$. In conclusion, the cost of an iteration is $O(N^{2.373})$.

\subsection{Principal hyper-parameters AND user-defined hyper-parameters}

Our proposed framework is applied to human-machine cooperation systems. Therefore, some terms depend on the user such as the failure mode definition which is defined by two factors: (i) how to determine whether two samples have a common concept; (ii) what the structure of a failure pattern is. In our experiments, we consider the case that the user defines an edge (common concept) by using the mutual $k_{nn}$-graph under the Euclidean distance on the embedding space. The connectivity criterion is maximally connected subgraphs (a.k.a.~connected components). With this indication, the user also provides two hyper-parameters $k_{nn}$ and $M$. The meaning of $k_{nn}$ and $M$ are mentioned in Appendix~\ref{sec:dataset}. From the algorithmic viewpoint, our approach depends mainly on one main hyper-parameter $\vartheta$. The parameter $\vartheta$ regulates the exploration-exploitation trade-off in the sampling procedure ($\vartheta = 0$ means pure exploitation, $\vartheta = 1$ means pure exploration). We experimented with five values of $\vartheta$ throughout the paper. 

\section{Proofs}
\subsection{Proofs of Proposition 6.1}
\begin{proof}[Proof of Proposition 6.1]
We first show that the value of $\delta$ should be upper-bounded by $\sqrt{N-1}$. To see this, note that $K(h_{\mc X}, h_{\mc Y})$ is a Gram matrix, so its diagonal elements are all ones, and the off-diagonal elements are in the range $(0, 1]$. We have an upper bound that:
\[
\| K(h_{\mc X}, h_{\mc Y}) - I_N \|_F \leq \sqrt{N(N-1)}.
\]
To ensure the existence of $h_{\mc X}, h_{\mc Y}$, the value of $\delta$ must fulfill:
\[
\delta \| I _N \|_F < \sqrt{N(N-1)}
 \implies 
\delta < \sqrt{N-1}.
\]
Next, we show that condition for $h_{\mc X}$ and $h_{\mc Y}$. Squaring both sides of~\eqref{eq:hyper-condition1} gives
\[
    \| K(h_{\mc X},  h_{\mc Y}) - I_N \|_F^2 \geq \delta^2 \| I _N \|_F^2 = \delta^2 N. 
\]
Because the diagonal elements of $K(h_{\mc X}, h_{\mc Y})$ are all ones, the above condition is equivalent to
\begin{align}
        \label{eq:hyper-condition2}
      &\sum_{i > j} \exp\big( -\frac{\| x_i - x_j \|_2^2}{h_{\mc X}^2} -\frac{\| \msa_{\hat{y}_i} - \msa_{\hat{y}_j} \|_2^2 + \| \covsa_{\hat{y}_i} - \covsa_{\hat{y}_j}\|_F^2}{h_{\mc Y}^2} \big)
      \geq \frac{\delta^2 N}{2}.
\end{align}
Using Jensen inequality for the exponential function, which is convex, we have the following lower bound:
\begin{align*}
&\frac{1}{{N \choose 2}}\sum_{i > j} \exp \big( -\frac{\| x_i - x_j \|_2^2}{h_{\mc X}^2} -\frac{\| \msa_{\hat{y_i}} - \msa_{\hat{y_j}} \|_2^2 + \| \covsa_{\hat{y_i}} - \covsa_{\hat{y_j}}\|_F^2}{ h_{\mc Y}^2} \big)
\\
& \qquad \geq \exp\big(-\frac{\sum_{i > j}\| x_i - x_j \|_2^2}{h_{\mc X}^2 {N  \choose 2}}  - \frac{\sum_{i > j}\| \msa_{\hat{y}_i} - \msa_{\hat{y}_j} \|_2^2 + \| \covsa_{\hat{y}_i} - \covsa_{\hat{y}_j}\|_F^2}{h_{\mc Y}^2 {N \choose 2}} \big).
\end{align*}
Therefore, if $h_{\mc X}$ and  $h_{\mc Y}$ satisfy 
\begin{align*}
\exp\big(-\frac{ \sum{i > j}\| x_i - x_j \|_2^2}{h_{\mc X}^2 {N  \choose 2}} - \frac{ \sum_{i > j}\| \msa_{\hat{y}_i} - \msa_{\hat{y}_j} \|_2^2 + \| \covsa_{\hat{y}_i} \covsa_{\hat{y}_j}\|_F^2}{h_{\mc Y}^2 {N \choose 2}}\big) 
\geq \frac{\delta^2}{N - 1} ,
\end{align*}
then they also satisfy the condition~\eqref{eq:hyper-condition2}. Defining the quantities $D_{\mc X}$ and $D_{\mc Y}$ as in statement of the proposition, we find that $h_{\mc X}$ and $h_{\mc Y}$ should satisfy
\[
\Leftrightarrow \frac{D_{\mc X}}{h_{\mc X}^2} + \frac{D_{\mc Y}}{h_{\mc Y}^2} \leq \ln{\frac{N-1}{\delta^2}}.
\]
This completes the proof.
\end{proof}

\subsection{Taylor Expansion for Value-of-Interest VoI}
\label{sec:taylor_expansion}
We first use a second-order Taylor expansion to approximate $f(X) = \VoI(X) = (1 + \exp(- g(X))^{-1}$ around the point $X=\mu$:
\begin{align*}
    f(X) 
    &= f(\mu) + (X - \mu)^\top \nabla f(\mu) + \frac{1}{2} (X - \mu)^\top \nabla^2 f(\mu) (X - \mu) + \mathcal{O}(\| \Delta_X \|^3) \\
    &= f(\mu) + (X - \mu)^\top \nabla f(\mu) + \frac{1}{2} \mathrm{Tr}[\nabla^2 f(\mu) (X - \mu) (X - \mu)^\top] + \mathcal{O}(\| \Delta_X \|^3).
\end{align*}
Moreover, we set $\mu$ as the expected value $\EE[X]$, and taking expectations on both sides of the above equation gives
\begin{align*}
    \EE[f(X)] 
    &= \EE\big[f(\mu)\big] + \EE\big[(X - \mu)^\top \nabla f(\mu)\big] + \frac{1}{2} \EE\big[\mathrm{Tr}[\nabla^2 f(\mu) (X - \mu) (X - \mu)^\top]\big] + \mathcal{O}(\| \Delta \|^3) \\
    &= f(\mu) + \frac{1}{2} \cov_{t, i}^* \nabla^2 f(\mu) + \mathcal{O}(\| \Delta \|^3),
\end{align*}
where the second equality follows from the relationship
\[
\EE\big[(X - \mu)^\top \nabla f(\mu)\big] = \EE\big[(X - \mu)\big]^\top \nabla f(\mu) = (\EE[X] - \mu)]^\top \nabla f(\mu) = 0,
\]
and from the definition of the covariance matrix
\[
\EE\big[(X - \mu) (X - \mu)^\top\big] = \cov_{t, i}^*.
\]
It now suffices to verify the expressions for $\alpha_i$ and $\beta_i$. Note that $\alpha_i = f(\mu) = (1 + \exp(- \mu))^{-1}$ and $\beta_i$ is the second-order derivative 
\begin{align*}    
    \beta_i &= \nabla^2 f(\mu) = \alpha_i(1-\alpha_i)(1-2\alpha_i) ,
\end{align*}
where the second equality follows from the property of the sigmoid function.

\section{SOCIAL IMPACT}

One important social impact of this research lies in its potential to improve the accuracy and reliability of machine learning classifiers. By identifying misclassification patterns, the framework enables the refinement and improvement of classifiers, reducing the likelihood of wrong predictions in various domains. This can have wide-ranging implications, such as improving the performance of automated systems in critical areas where accurate classification is of utmost importance like healthcare diagnosis~\citep{shaban2021guest, rudin2018optimized, albahri2023systematic}, or autonomous vehicles~\citep{glomsrud2019trustworthy, wagner2015philosophy}.

Another significant social impact of this research is its potential to address biases and fairness issues in machine learning systems~\citep{caton2020fairness, mehrabi2021survey, pessach2022review}. By identifying misclassification patterns, the framework can shed light on potential biases in the data or algorithmic models. This knowledge is crucial for developing fairer and more equitable machine learning systems which are obligatory for bringing machine learning models to practical implementations.

Moreover, the collaborative nature of the framework promotes human-machine interaction, fostering a symbiotic relationship that combines human expertise and algorithmic capabilities. This approach not only empowers human annotators by involving them in the decision-making process but also allows them to contribute their domain knowledge and intuition~\citep{wu2022survey, xin2018accelerating}.

\end{document}